%% file: main.tex
\documentclass[sn-mathphys]{sn-jnl}% Math and Physical Sciences Reference Style
\jyear{2021}%

\usepackage{subfig}
\usepackage{soul}
\usepackage{xcolor}
\usepackage{hyperref}
\hypersetup{colorlinks=true,linkcolor=red,urlcolor=red}

\theoremstyle{thmstyleone}%
\theoremstyle{thmstyletwo}%

\theoremstyle{thmstylethree}%

\newcommand{\acronym}{FedAR}

\begin{document}

%\title[Semi-Supervised Federated Learning for Human Activity Recognition]{\acronym: Personalized Semi-Supervised Federated Learning for Human Activity Recognition}
%\title[Semi-Supervised Federated Learning for Human Activity Recognition]{Semi-Supervised and Personalized Federated Activity Recognition Based on Active Learning and Label Propagation}
\title{Personalized Semi-Supervised Federated Learning for Human Activity Recognition}

\author{\fnm{Riccardo} \sur{Presotto}}\email{riccardo.presotto@unimi.it}

\author*{\fnm{Gabriele} \sur{Civitarese}}\email{gabriele.civitarese@unimi.it}

\author{\fnm{Claudio} \sur{Bettini}}\email{claudio.bettini@unimi.it}

\affil{\orgdiv{Dept. of Computer Science}, \orgname{University of Milan}, \city{Milan}, \country{Italy}}


\abstract{%226 words -> maximum 250!

%Sensor-based Human Activity Recognition (HAR) is a widely explored research area. The most effective data-driven methods for human activities prediction are based on supervised machine learning applied to the continuous stream of sensors data. 
%
One of the major open problems in sensor-based Human Activity Recognition (HAR) is the scarcity of labeled data.
%, these methods perform well on restricted sets of activities in domains for which there is a 
%fully labeled dataset. It is still a challenge to cope with the intra- and inter-variability of activity execution among different subjects that characterizes a large scale real world deployment. 
%Even though existing supervised solutions exhibit satisfying results on public labeled datasets, the deployment of those approaches in real-word scenarios is still challenging. Indeed, the execution of human activities is characterized by intra- and inter-variability among different subjects. Hence, a large training set should be used to achieve solutions that generalize over unseen users and that can also be fine-tuned on each subject. However,
%
Among the many solutions to address this challenge, semi-supervised learning approaches
represent a promising direction.
%challenge of acquiring the large amount of labeled data 
%that is necessary in realistic settings.
However, their centralised architecture incurs in the scalability and privacy problems that arise when the process involves a large number of users.
Federated Learning (FL) is a promising paradigm to address these problems. However, the FL methods that have been proposed for HAR assume that the participating users can always obtain labels to train their local models (i.e., they assume a fully supervised setting). In this work, we propose \acronym: a novel hybrid method for HAR that combines semi-supervised and federated learning to take advantage of the strengths of both approaches. \acronym\ combines active learning and label propagation to semi-automatically annotate the local streams of unlabeled sensor data, and it relies on FL to build a global activity model in a scalable and privacy-aware fashion. \acronym\ also includes a transfer learning strategy to fine-tune the global model on each user. 
%We designed a novel evaluation methodology to assess the personalization and generalization capabilities of our approach. 
We evaluated our method on two public datasets, showing that \acronym\ reaches recognition rates and personalization capabilities similar to state-of-the-art 
%locally 
FL supervised approaches. As a major advantage, \acronym\ only requires a very limited number of annotated data to populate a pre-trained model and a small number of active learning questions that quickly decrease while using the system, leading to an effective and scalable solution for the data scarcity problem of HAR. 
%Our results also show that both the generalization and personalization capabilities of \acronym~ continuously improve over time.
}




\maketitle

\begin{center}
{\small
\textcolor{red}{
\textbf{Note:} This paper has been published in \textit{Personal and Ubiquitous Computing}.\\
Please cite as: Presotto, R., Civitarese, G., \& Bettini, C. (2022). Semi-supervised and personalized federated activity recognition based on active learning and label propagation.
\textit{Personal and Ubiquitous Computing}, 26(5), 1281–1298. \href{https://doi.org/10.1007/s00779-021-01627-z}{https://doi.org/10.1007/s00779-021-01627-z}
}
}
\end{center}

\section{Introduction}
\input{sections/introduction}

\section{Related Work}
\input{sections/related}

\section{Overview of \acronym}
\input{sections/architecture}

\section{\acronym\ under the hood}
\input{sections/method}

\section{Experimental evaluation}
\input{sections/experiments}

\section{Discussion}
\input{sections/discussion}
\section{Conclusion and future work}
\input{sections/conclusion}


\section*{Conflict of interest statement}
On behalf of all authors, the corresponding author states that there is no conflict of interest.

\bibliography{references}% common bib file


\end{document}

%% file: sections/introduction.tex
%The evolution of mobile computing and sensor technologies in the last decades made it possible to develop intelligent applications that continuously monitor our daily activities to enable context-aware services~\cite{chen2012sensor}.
%Off-the-shelf mobile and wearable devices that are pervasively adopted in our everyday life can be used to track physical movements and hence to constantly monitor daily activities. 
%c
%In the literature, the majority of approaches for Human Activity Recognition (HAR) based on mobile and wearable devices rely on supervised learning methods to infer activities from the continuous stream of data generated by the sensors embedded in those devices~\cite{lara2013survey}.
The majority of approaches for Human Activity Recognition (HAR) based on data continuously acquired from mobile and wearable devices rely on supervised learning methods~\cite{lara2013survey}.

While supervised learning leads to high recognition rates, collecting a 
%significant 
sufficiently representative
amount of labeled data to train the recognition model is often a real challenge~\cite{CookFK13}. 
%costly, time-consuming, intrusive and hence prohibitive~\cite{CookFK13}.
For instance, data annotation can be performed directly by the monitored subject while performing activities (self-annotation). However, this approach is very obtrusive and error-prone. Alternatively, external observers can annotate the activity execution of a subject (in real-time or by semi-automatic video annotation), but these methods are time-consuming and privacy-intrusive.

Among the solutions that have been proposed to tackle the labeled data scarcity problem, semi-supervised learning represent a promising research direction that has been explored in the last few years~\cite{abdallah2018activity}. Semi-supervised methods only use a small amount of labeled data to initialize the recognition model, which is continuously updated taking advantage of unlabeled data. %
%The most common semi-supervised approaches are label propagation ~\cite{longstaff2010improving}, co-learning~\cite{lee2014activity}, and active learning~\cite{ miu2015bootstrapping, abdallah2015adaptive, hossain2017active}.
%
However, there are still several challenges that limit the deployment of these methods in realistic scenarios. Indeed, even though semi-supervised approaches mitigate the data scarcity problem, they do not consider the scalability and privacy issues that arise in training a real-world recognition model that includes data from a large number of different users. From the scalability point of view, the computational effort that is required to train a global model significantly grows as the number of users increases. Considering privacy aspects,  activity data may reveal sensitive information, like the daily behavior of a subject and her habits~\cite{bettini2015privacy}.
%A continuously evolving activity model based on data from a large number of users is desirable for real-world HAR. 
Accurate HAR also requires a certain amount of personalization on the end users~\cite{weiss2012impact}.
%usually require to centralising the collected data on the machine where resides the machine learning model. This strategy may pose a considerable threat to user privacy, especially in the case of activity recognition where the required training data could reveal sensitive information (e.g., health conditions or habits) \cite{voigt2017eu, samarati2014data, bettini2015privacy}.

In 2016, Google introduced the Federated Learning (FL) framework \cite{mcmahan2017communication}. In the FL paradigm, the model training task is distributed over a multitude of nodes (e.g., mobile devices). 
Each node uses its own labeled data to train a local model. The resulting model parameters of each participating node are forwarded to a server that is in charge of aggregating them. Finally, the server shares the aggregated parameters to the participating nodes.
FL is a promising direction to make activity recognition scalable for a large number of users. Moreover, FL 
%protects users' privacy, 
mitigates the privacy problem since only model parameters, and not actual data, are shared with the server, and privacy-preserving mechanisms (e.g., Secure MultiParty Computation, Differential Privacy) are used when aggregating parameters~\cite{yang2019federated}.

FL has been recently applied to HAR showing that it can reach an accuracy very close to centralised methods~\cite{chen2020fedhealth}. 
However, all existing solutions assume that each node has complete availability of labeled sensor data. 
This is actually the general setting of existing works based on FL, that in the literature has been primarily considered for fully supervised learning tasks~\cite{kairouz2019advances}.
While this assumption may be valid for some applications (e.g., the Google approach for keyboard suggestions improvement relies on labeled data implicitly provided by users when typing or confirming suggestions~\cite{hard2018federated}), it is not realistic for applications like HAR where labeled data availability is significantly limited. Extending FL to semi-supervised learning is one of the open challenges in this area~\cite{kairouz2019advances}.

In this work, we propose \acronym: a hybrid semi-supervised and FL framework that enables personalised privacy-aware and scalable HAR based on mobile and wearable devices. Different from the majority of the existing solutions, \acronym\ considers a limited availability of labeled data. In particular, \acronym\ combines active learning and label propagation to provide labels to a large amount of unlabeled data. Newly labeled data are periodically used by each node to perform local training, thus obtaining the model parameters that are then transmitted to the server that aggregates them using Secure Multiparty Computation. \acronym\ also relies on transfer learning to fine-tune the global model for each user, while generating a global model that generalizes over unseen users.

Considering the limitations of existing evaluation methodologies for FL applied to HAR~\cite{ek2020evaluation}, we designed a novel evaluation methodology to robustly assess both the generalization and the personalization capabilities of our approach. The results of our experimental evaluation on two publicly available datasets show that \acronym\ reaches recognition rates close to state-of-the-art solutions that assume the complete availability of labeled data. Moreover, both the generalization and the personalization capabilities of \acronym\ keep increasing over time. Last but not least, the amount of triggered active learning questions is small and acceptable for a real-word deployment.

To the best of our knowledge, \acronym\ is the first FL framework for HAR that tackles the data scarcity problem while considering personalization. Hence, we believe that \acronym\ is a significant step towards realistic deployments of HAR systems based on FL. 

%Even if we evaluated \acronym~ only considering as target application activity recognition based on mobile and wearable devices, our method can also be applied to other HAR applications (e.g., HAR in smart-homes with environmental sensors) and more generally to other human-centered domains characterized by low availability of labeled data~\footnote{We better characterize the generality of our approach in Section~\ref{subsec:generality}.}.

In summary, the contributions of this work are the following:
\begin{itemize}
    \item We present \acronym, a novel hybrid approach that combines federated, semi-supervised, and transfer learning to tackle the data scarcity problem for real-world personalized HAR.
    \item We propose a novel strategy to reliably evaluate the evolution of the personalization and generalization capabilities of \acronym\ over time.
    \item An extensive evaluation on public datasets shows that \acronym\ reaches similar recognition rates with respect to well-known approaches that assume high-availability of labeled data. At the same time, \acronym\ triggers a small number of active learning questions that quickly decreases while using the system.
\end{itemize}

%% file: sections/related.tex
\label{sec:related}

%aggiornare su data scarcity problem più in generale
\subsection{Labeled data scarcity in HAR}
Considering HAR based on data collected from mobile devices' inertial sensors, the majority of  approaches rely on supervised machine learning~\cite{kwapisz2011activity, gyorbiro2009activity, sun2010activity, bao2004activity, BullingBS14}. However, these approaches need a significant amount of labeled data to train the classifier. Indeed, different users may perform the same activities in very different ways, but also distinct activities may be associated with similar motion patterns. The annotation task is costly, time-consuming, intrusive, and hence prohibitive on a large scale~\cite{CookFK13}.
In the following, we summarize the main methodologies that have been proposed in the literature to mitigate this problem.

Unsupervised approaches have been proposed to derive activity clusters from unlabeled sensor data~\cite{kwon2014unsupervised}. Those approaches still need annotations to reliably associate an activity label to each cluster. Since distinct human activities often share similar sensor patterns, purely unsupervised data-driven approaches for activity recognition are still a challenge considering real-world scenarios.

Some research efforts focused on knowledge-based approaches based on logical formalisms, especially targeting smart-home environments~\cite{chen2009ontology,civitarese2019polaris}. These approaches usually rely on ontologies to represent the common-sense relationships between activities and sensed data. One of the main issues of knowledge-based approaches is their inadequacy to model the intrinsic uncertainty of sensor-based systems and the large variety of activity execution modalities. %Indeed, logic-based models for HAR are generally hand-crafted by domain experts and knowledge engineers.

Data augmentation is a more popular solution adopted in the literature to mitigate the data scarcity problem, especially considering imbalanced datasets~\cite{chawla2002smote,rashid2019times}. In these approaches, the available labeled data are slightly perturbed to generate new labeled samples. With respect to our method, data augmentation is an orthogonal approach that could be integrated to further increase the amount of labeled data. %Indeed, in future work we plan to investigate if data augmentation can improve the results of our method. 
Recently, data augmentation in HAR has also been tackled taking advantage of GAN models to generate synthetic data more realistic than the ones obtained by the above-mentioned approaches~\cite{wang2018sensorygans,chan2020unified}. However, GANs require to be trained with a significant amount of data. 

Many transfer learning approaches have been applied to HAR to fine-tune models learned from a source domain with available labeled data to a target domain with low-availability of labeled data~\cite{cook2013transfer,wang2018deep, sanabria2021unsupervised, soleimani2021cross}. \acronym~ relies on transfer learning to fine-tune the personal local model taking advantage of the global model trained by all the participating devices. 

An effective method to tackle data scarcity for HAR when the feature space is homogeneous (like in \acronym) is semi-supervised learning~\cite{abdallah2018activity, stikic2008exploring, guan2007activity, longstaff2010improving}. Semi-supervised methods only use a restricted labeled dataset to initialize the activity model. Then, a significant amount of unlabeled data is semi-automatically annotated. The most common semi-supervised approaches for HAR are self-learning ~\cite{longstaff2010improving}, label propagation~\cite{stikic2009multi}, co-learning~\cite{lee2014activity}, and active learning~\cite{miu2015bootstrapping, abdallah2015adaptive, hossain2017active}. 
Active learning has also been adopted in HAR to handle the class imbalance problem~\cite{nguyen2018dealing}.
Hybrid solutions based on semi-supervised learning and knowledge-based reasoning have been proposed in~\cite{bettini2020caviar}.
%~\cite{bettini2020caviar,abdallah2018activity, stikic2008exploring, guan2007activity, longstaff2010improving}.
Existing semi-supervised solutions do not consider the scalability problems related to building a recognition model with a large number of users for real-world deployments.
%require usually centralising the collected data on the machine where resides the machine learning model. 
Moreover, the data required to build such collaborative models is sensitive, as it could reveal private information about the users (e.g., user health condition and habits)~\cite{voigt2017eu, samarati2014data, bettini2015privacy}.

\subsection{Federated Learning for HAR}
Recently, the FL paradigm has been proposed to distribute model training over a multitude of nodes~\cite{yang2019federated,konevcny2016federated,bonawitz2017practical,mcmahan2017communication, kairouz2019advances, damaskinos2020fleet}. A recent survey divides FL methods in three categories: horizontal, vertical, and transfer FL~\cite{yang2019federated}. \acronym~ is a horizontal FL method: the participating mobile devices share the same feature space, but they have a different space in samples (i.e., each device considers data for a specific user). Among the required characteristics of FL approaches, the personalization of the global model on each client plays a major role~\cite{fallah2020personalized}
%Each client uses its labeled data to perform the training task locally, and only the learned model parameters are forwarded to a server.
%Hence, the sensitive data used to train the model is never shared with other entities. The server is in charge of aggregating model parameters from multiple clients. Finally, each participating client receives the updated parameters from the server. This process is repeated periodically. 
%(e.g., each night when devices are idle and charging). 
%Since FL regarding privacy concerns and the possibility of distributing the learning process on a myriad of devices, in recent years many of its applications have been proposed in the field of mobile devices based activity recognition [mettere citazioni].
% spiegare uno ad uno cosa fanno i lavori proposti

FL has been previously applied to mobile/wearable HAR to distribute the training of the activity recognition model among the participating devices~\cite{ek2021federated,sozinov2018human,wu2020personalized,zhao2020semi,chen2020fedhealth,wu2020fedhome,ek2020evaluation}. In this area, recent works also proposed to learn the global model in a decentralized fashion~\cite{lee2021opportunistic}.
Existing works show that FL solutions for HAR reach recognition accuracy similar to standard centralized models~\cite{sozinov2018human}. Moreover, since personalization is an important aspect for HAR~\cite{weiss2012impact}, existing works also show that applying transfer learning strategies to fine-tune the global model on each client leads to a significantly improved recognition rate~\cite{wu2020personalized,chen2020fedhealth}.
One of the major drawbacks of these solutions is that they assume high availability of labeled data, hence considering a fully supervised setting.
%This is actually a general problem of FL that, in the literature, has been mainly proposed for supervised learning tasks where labels are naturally available on each client~\cite{kairouz2019advances}.

The combination of federated and active learning has been recently proposed for Intrusion Detection Systems~\cite{kelli2021ids}.
However, semi-supervised federated learning solutions for HAR have been only partially explored. The existing works mainly focus on unsupervised methods to collaboratively learn (based on the FL setting) a robust feature representation from the unlabeled stream of sensor data. The global feature representation is then used to build activity classifiers using a limited amount of labeled data. For instance, the work in~\cite{zhao2020semi} proposes an approach based on autoencoders, while the work in~\cite{saeed2020federated} is based on self-supervised learning. However, those works do not consider model personalization and they do not propose approaches to continuously obtain new labeled data from each user.
Nonetheless, we believe that those works focus on a very important orthogonal problem with respect to the one addressed by \acronym. Indeed, feature learning from unlabeled data could be integrated in  \acronym\ to further reduce the amount of active learning questions and to improve the recognition rate.
Recently, the work in~\cite{yu2021fedhar} proposes a solution to build the global model by aggregating the local models' gradients from a small number of clients with labeled data and a large number of clients with unlabeled data. This method is based on a semi-supervised loss that relies on a novel unsupervised gradients aggregation strategy. Differently from this work, we do not assume the existence of clients with full availability of labeled data, and we also propose a practical solution to continuously improve the global model thanks to active learning and label propagation.

%Differently from existing works, \acronym~ considers a setting where only a limited amount of labeled data is available: a small training set that is only used to initialize the global model.
%Then, each client uses a combination of active learning (in real-time) and label propagation to associate labels to the continuous stream of unlabeled data. 

A common limitation in the literature is the methodology adopted to evaluate FL for HAR applications~\cite{ek2020evaluation,chen2020fedhealth,wu2020fedhome}. Indeed, none of the proposed methodologies truly assess the generality of the global model over users whose data have never been used for training. Moreover, only one iteration of the FL process is evaluated, while in a realistic deployment this process is repeated periodically with different data. In this work, we propose an evaluation methodology that overcomes the above-mentioned issues (see Section~\ref{subsec:evaluationmethodology}).

%elogia articolo di flora e dire in modo gentile cosa facciamo qui e cosa non fanno loro. 

%% file: sections/architecture.tex
In this section, we describe the overall \acronym\ framework at a high-level.

\subsection{Overall architecture}
\begin{figure}[h!]
	\centering
	\includegraphics[scale=0.3]{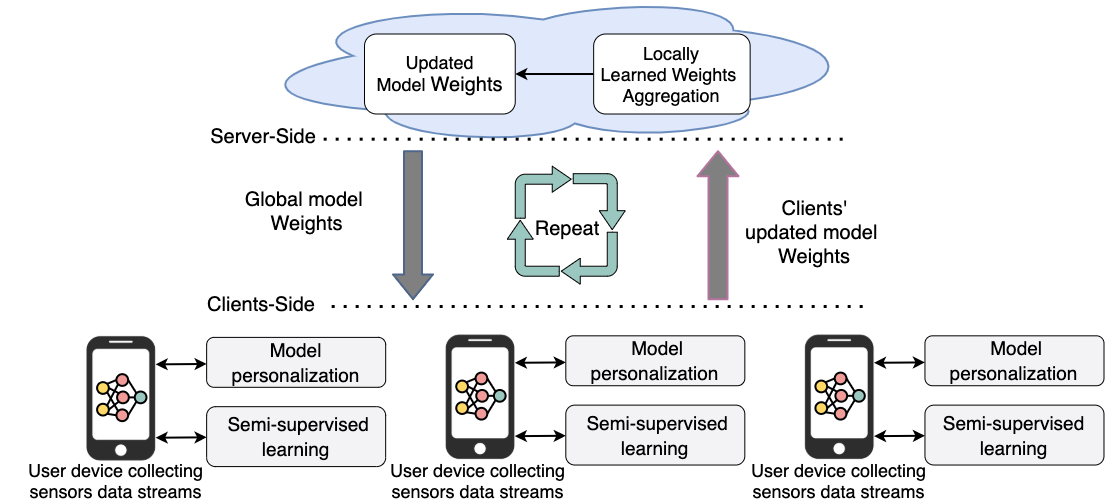}
	\caption{Overall architecture of \acronym.}
	\label{fig:global_data_flow}
\end{figure}

The overall architecture of \acronym\ is depicted in Figure~\ref{fig:global_data_flow}.
For the sake of this work and without loss of generality, we 
%contextualize 
illustrate \acronym~ applied to physical activity recognition based on inertial sensors data collected from personal mobile devices.

%\subsubsection{Federated Learning}
Following the FL paradigm, the actors of \acronym\ are a server and a set of clients that cooperate to periodically compute the weights of a global activity recognition model. 
In order to address 
the labeled data scarcity problem, \acronym\ initializes the global model in an offline phase with a limited amount of labeled data, while each client implements a semi-supervised learning strategy (i.e., a combination of active learning and label propagation) to semi-automatically label a portion of the unlabeled sensor data stream. An overview of our semi-supervised strategy is described in Section~\ref{subsec:semioverview}. 

Periodically (e.g., every night), the server starts a process to update the weights of the global model. Each client uses its available labeled data to train its local model. The resulting local weights are transmitted to the server, that aggregates them with the ones from the other clients to obtain a new version of the global model. Finally, the new version of the global model weights are transmitted to each client.
Since different users may perform activities in very different ways, a model personalization module on each client is in charge of fine-tuning
%it. The personalized 
the updated global model weights on the specific user. A more detailed overview on the global model update and personalization is described in Section~\ref{subsec:globalpersoverview}.

\subsection{Local models}

One of the strengths of \acronym\ is that it is designed considering both personalization and generalization aspects. Personalization is crucial for the local models to recognize the activities of each user more accurately. On the other hand, generalization is a desirable property for the global model. Indeed, 
%a portion 
some 
%of the 
participating users may not wish to collect labeled data (not even a small amount), or may have 
%physical 
devices 
%that are 
not adequate to perform local training. Those users are not able to actively contribute to the federated learning process, and their clients would directly use the last version of the global model for activity classification. 

In order to guarantee both personalization and generalization, in \acronym, each client stores two distinct instances of the activity model.  
The former is called \textit{Shareable Model}, and it is the one used for federated learning. 
In order to personalize the activity model on each user, a straightforward solution would be to fine-tune the \textit{Shareable Model} with transfer learning approaches~\cite{arivazhagan2019federated}. However, recent studies shows that a global model built by aggregating the weights of fine-tuned models exhibit poor generalization capabilities on external users~\cite{ek2020evaluation}.
In order to overcome this problem, in \acronym, at the end of each global model update the clients that actively contribute to the federated learning process create a copy of the \textit{Shareable Model} that is called \textit{Personalized Model}. The \textit{Personalized Model} is fine-tuned on the specific user and it is used for activity classification. 
Besides improving generalization, an advantage of keeping private the weights of the \textit{Personalized Local Model} is a positive impact on privacy protection~\cite{nasr2019comprehensive}.

\subsection{Semi-supervised data labeling and classification}
\label{subsec:semioverview}

\begin{figure}[h!]
	\centering
	\includegraphics[scale=0.4]{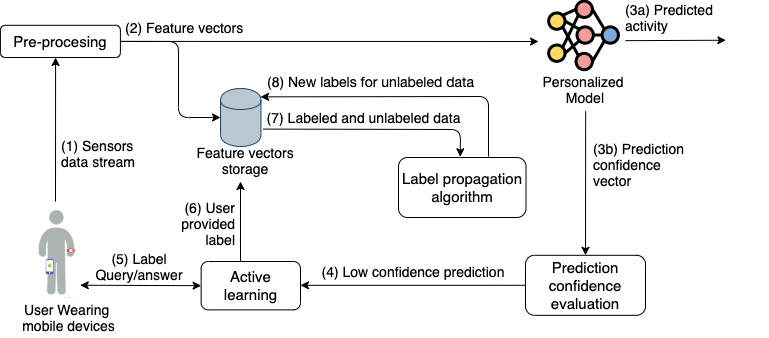}
	\caption{Semi-supervised data labeling and classification data flow}
	\label{fig:data_labelling}
\end{figure}

Figure~\ref{fig:data_labelling} depicts the semi-supervised data labeling and classification flow of \acronym.

%When a node receives updated global model weights from the server, the \textsc{Model Personalization} module applies transfer learning methods to fine-tune the global model on the user, thus updating the
Each client in \acronym\ uses the \textit{Personalized Model} to classify activities in real-time on the continuous stream of unlabeled pre-processed sensor data. Before classification, each unlabeled data sample is stored in the \textit{Feature Vectors Storage}. This storage collects both unlabeled and labeled data samples. After classification, if the confidence on the current prediction is below a threshold, an active learning process is started, and the system asks the user about the activity that she was actually performing. The feedback from the user is then associated with the corresponding feature vector in the \textit{Feature Vectors Storage}. 
Active learning makes it possible to assign a label to those informative data points that can effectively improve the local model.
For the sake of usability, the number of active learning queries should be low, since they may bother the user during activity execution. For this reason, \acronym\ also periodically applies a Label Propagation algorithm to spread the labels acquired through active learning to a larger number of unlabeled data points. The advantage of label propagation is to further improve the recognition rate by training the classifier with a significant amount of labeled data samples and, at the same time, to reduce the number of needed active leaning queries over time.
%The classification and active learning process is also described as pseudo-code in Algorithm~\ref{alg:clientactive}.

%\begin{figure}[h!]
%\centering
%    \subfloat[Classification and active learning]{{\includegraphics[width=6.9cm]{img/data_labelling.png} 
%	\label{fig:data_labelling}
%	}}
%	\qquad
%		\subfloat[Labels propagation and local model training]{{\includegraphics[width=6.7cm]{img/local_training.png} 
%	\label{fig:local_training}
%	}}
%	\caption{An overview of the \acronym~ framework.}
%\label{fig:architectures}
%\end{figure}

\subsection{Global model update and personalizazion}
\label{subsec:globalpersoverview}

\begin{figure}[h!]
	\centering
	\includegraphics[scale=0.35]{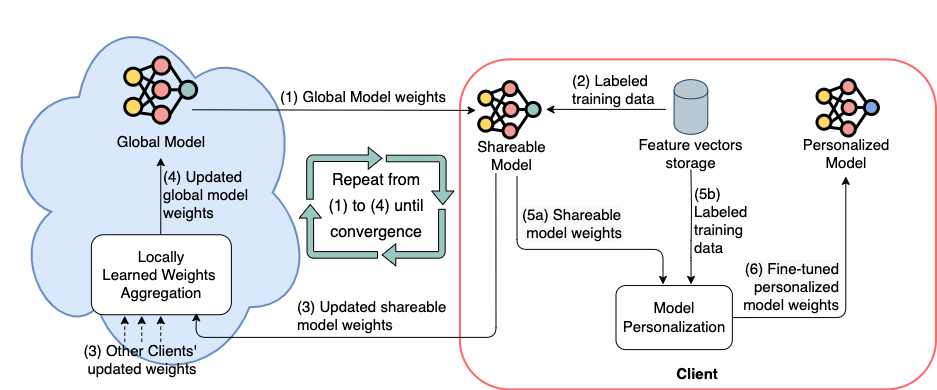}
	\caption{Local models training and personalized model update}
	\label{fig:local_training}
\end{figure}

Periodically (e.g., every night) the server asks to the participating clients to update the global model.
This process is depicted in Figure~\ref{fig:local_training}.

First, each client replaces its \textit{Shareable Model} with the current version of the \textit{Global Model}. Then, the labeled data in the \textit{Feature Vectors Storage} are used to perform local training of the \textit{Shareable Model}. After training, the updated \textit{Shareable Model} weights are then forwarded to the server, that is in charge of aggregating the weights from all the clients to generate a new version of the global model. These steps are repeated until convergence of the global model. At the end of this process, the \textit{Shareable Model} of each client is replaced with the last stable version of the \textit{Global Model}.
%Once the global model is stable, the participating clients  \textit{Shareable Models}.

Then, the \textit{Model Personalization} module generates a copy of the \textit{Shareable Model} that is called \textit{Personalized Model}, that is fine-tuned using the \textit{Feature Vectors Storage}. The result of this process is a \textit{Personalized Model} that takes advantage of the high-level features of the \textit{Global Model} as well as the personalized aspects of the specific user.
%Note that the resulting model is not influenced by the \textit{Global Model} training process.
%Finally, the \textit{Shareable Model} and the \textit{Personalized Model} are combined by the \textit{Model Personalization} module to generate a new \textit{Personalized  Model} that 
%As we previously mentioned, only the weights of the \textit{Local Model} are actually transmitted to the server during the FL process to update the global model. We will describe our federated learning strategy in more details in Section~\ref{subsec:federated}.
%The \textit{pre-training dataset} is also used to initialize the Label Propagation model that is distributed to all the participating devices. More details about Label Propagation will be presented in Section~\ref{subsec:label_spreading}.

%% file: sections/method.tex
%In this section we describe in detail the methodology adopted by \acronym. First, we describe the overall architecture of \acronym. Then, we describe how the global model is initialized and 
%\subsection{Overview of \acronym}
%In the following, we introduce the overall architecture of \acronym. 
%that consists of a novel implementation of a FL framework in the field of semi-supervised activity recognition.

In this section, we describe in detail the algorithms of \acronym.

\subsection{The activity model}
\label{subsec:model}

Since we consider a setting with limited availability of labeled data, activity models that automatically learn features from raw data are not effective in \acronym. Indeed, based on our experiments that we describe in Section~\ref{subsubsec:mlp_vs_cnn}, CNN models reach significantly lower recognition rates in \acronym\ due to the high complexity of learning reliable features from limited labeled data.
For this reason, in \acronym, the activity classification model is based on a fully-connected deep learning model, and the input is a vector of handcrafted features. In particular, we choose features that proved to be effective for HAR~\cite{bettini2020caviar}. Recent studies in the HAR domain demonstrate that a good choice of handcrafted features and fully connected models can lead to recognition rates comparable to the ones of state-of-the-art CNN models~\cite{cruciani2019comparing}.

%
%the restricted data availability would lead to an unreliable feature representation, negatively impacting the recognition rate during classification.
%that would not scale on the large amount of users participating to the FL process.
In particular, for each axis of each inertial sensor, we consider the following features: \textit{average, variance, standard deviation, median, mean squared error, kurtosis, symmetry, zero-crossing rate, number of peaks, energy, and difference between maximum and minimum}.
These features are extracted from fixed-length temporal windows of sensor data of $w$ seconds. Before feature extraction, we apply a median filter on each temporal window to reduce noise in sensor measurements. After feature extraction, we apply standardization as a feature scaling technique.

%While recent methods propose to automatically extract features from unlabeled sensor data streams in HAR federated learning settings~\cite{saeed2020federated}, we believe that feature learning represents an orthogonal problem.

\subsection{Initialization of the global model}
\label{subsec:pre_train}

At the very beginning, the participating clients in \acronym\ need a pre-trained global model to infer labels on unlabeled data. However, in this work we assume limited availability of labeled data. 

\begin{figure}[h!]
	\centering
	\includegraphics[scale=0.4]{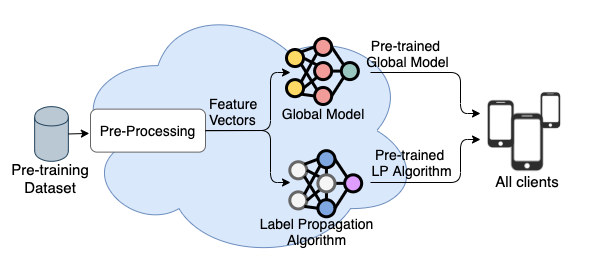}
	\caption{Initialization of the global model in \acronym.}
	\label{fig:pre_train}
\end{figure}

Hence, \acronym\ initializes the global model using a restricted annotated dataset (we will call it \textit{pre-training dataset} in the following)~\footnote{Note that, considering out target application, a labeled dataset is a collection of timestamped inertial sensors data acquired from mobile/wearable devices during activity execution. Examples of such sensors are accelerometer, gyroscope, and magnetometer. The labels are annotated time intervals that indicate the time-span of each performed activity.}. The \textit{pre-training dataset} is also used to initialize label propagation algorithm. In realistic settings, the \textit{pre-training dataset} can be, for example, a combination of publicly available datasets, or a small training set specifically collected by a restricted number of volunteers.
Figure~\ref{fig:pre_train} summarizes the initialization mechanism of \acronym.

\subsection{The Federated Learning strategy of \acronym}
\label{subsec:federated}

In the following, we describe the FL process to update the global and local models.
%\acronym~ takes advantage of FL to distribute the training of the global activity model over a large number of devices.
%Each participant contributes to training the global model with a certain amount of labeled data. %We will explain in %Section~\ref{subsec:class_and_lab} how labeled data are actually collected thanks to semi-supervised learning.
%
Periodically (e.g., each night) the server starts a global model update process. The devices that are available to perform computation (e.g, the ones idle and charging) inform the server that they are eligible to take part in the FL process. Afterwards, the server executes several communication rounds to update the weights of the global model. 

A \textit{communication round} consists of the following steps:
\begin{itemize}
    \item The server sends the latest version of the global weights to a fraction of the eligible devices %(for the sake of scalability)
    \item Each device uses the labeled data in the \textit{Feature Vectors Storage} to train the \textit{Shareable Model}
    \item When local training is completed, each device sends the new weights of the \textit{Shareable Model} to the server
    \item The server aggregates the local weights to compute the new global weights
\end{itemize}

The communication rounds are repeated until the global model converges. 
%We use a fixed number obtained by experiments on the left out users
Then, the new weights are transmitted to each participating device including the ones that did not actively contribute to the communication rounds. 

%The \acronym~ component on the device will personalize the model (as explained in Section~\ref{sec:personalization}), and use it for local activity classification.
%As proposed in FedAVG~\cite{mcmahan2017communication}, 
The server updates the global model weights by executing a weighted average of the locally learned model weights provided from clients. 
%\acronym~ adopts this aggregation strategy as it is the one most commonly used in the FL approaches proposed in the literature. 
Since the local weights may reveal private information, the aggregation is performed using the Secure Multiparty Computation approach presented in~\cite{bonawitz2017practical}. 
The pseudo-code of the server-side federated learning process is described in Algorithm~\ref{alg:serverside}, while the client-side in Algorithm~\ref{alg:modelupdate}.

\begin{algorithm}
\caption{Server side - Federated global model}\label{alg:serverside}
%\hspace*{\algorithmicindent} \textbf{Input} \\
%\hspace*{\algorithmicindent} \textbf{Output} 
\begin{algorithmic}[1]
\State $PT \Leftarrow $ pre-training set 
\State Initialize global model $w^G$ with $PT$
\State $d \Leftarrow $ participating devices
\For{each periodic update (e.g., every night)}

  \For{each communication round}
  \State ask for eligibility to each device in $d$
  \State $ed \Leftarrow$ eligible devices
  \State $ed' \Leftarrow$ $k$ devices randomly sampled from $ed$
  \State send $w^G$ to each device in $ed'$
  \State aggregate updated models' weights received from devices in $ed'$ with SMC~\cite{bonawitz2017practical}
  \EndFor
       
\EndFor
\end{algorithmic}
\end{algorithm}

\begin{algorithm}
\caption{\acronym\ - Client side - Model update}\label{alg:modelupdate}
%\hspace*{\algorithmicindent} \textbf{Input} \\
%\hspace*{\algorithmicindent} \textbf{Output} 
\begin{algorithmic}[1]
%\State Update User Device Storage by Applying Label propagation algorithm (Section~\ref{subsec:label_spreading})
\State $pm \Leftarrow $ Personalized Model
\State $sm \Leftarrow $  Shareable Model

\State Update the \textit{Feature Vectors Storage} using the Label Propagation algorithm in Section~\ref{subsec:label_spreading}.

\For{each communication round $i$}
 \State train $sm$ using labeled data in the Feature Vector Storage
 \State send $sm$ to the server
 \State receive updated global model $w_i^G$
 \State $sm \Leftarrow$ $w_i^G$
 % capure se linea sopra è pm o sm
\EndFor
\State $pm \Leftarrow sm$
\State fine-tune $pm$ using the transfer learning method described in Section~\ref{sec:personalization}

%\State train $pm$ using labeled data in the Feature Vector Storage
% \State $pm \Leftarrow$ combine $sm$ and $pm$ with the transfer learning in Section~\ref{sec:personalization} for personalization.

\end{algorithmic}
\end{algorithm}

\subsection{Model Personalization}
\label{sec:personalization}
\acronym\ adopts a transfer learning strategy to fine-tune the \textit{Personalized Model} on each user. The intuition behind the personalization mechanism is that the last layers of the neural network (i.e., the ones closer to the output) encode personal characteristics of activity execution, while the remaining layers encode more general features that are common between different users~\cite{yosinski2014transferable}.
%   A similar mechanism has been also used in a recent FL approach to HAR~\cite{wu2020personalized}.
%This is indeed the approach used by \acronym~ for personalization.
\begin{figure}[h!]
	\centering
	\includegraphics[scale=0.3]{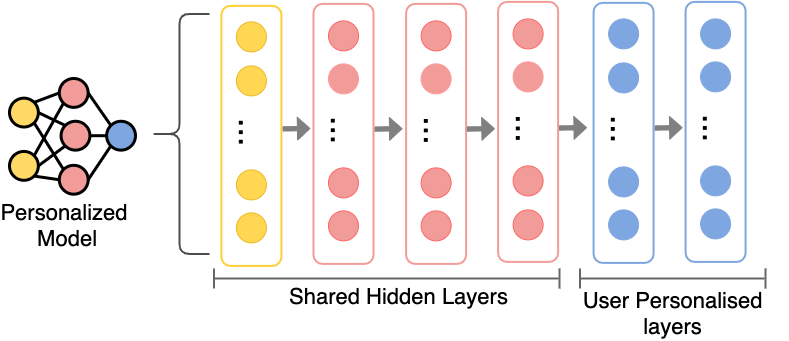}
	\caption{Shared and Personal Layers.}
	\label{fig:personalization_layers}
\end{figure}
As depicted in Figure~\ref{fig:personalization_layers}, we refer to the last $l$ layers of the neural network as the \textit{User Personalized Layers}, while we refer to the remaining ones as \textit{Shared Hidden Layers}. 
In \acronym, when the update of the global model is complete, each client creates the  \textit{Personalized Model} as a copy of the \textit{Shareable Model}.
In order to fine-tune the \textit{Personalized Model} on each user, the \textit{Shared Hidden Layers} are freezed, and the \textit{Feature Vector Storage} is used to train the \textit{User Personalized Layers}.

%of the \textit{Shared Hidden Layers} are replaced with the ones of the \textit{Shareable Model} (i.e., the last stable version of the global model), while the \textit{User Personalized Layers} are not updated. 
%As we previously mentioned, the weights of the \textit{Personalized Local Model} are not used for the FL process, and hence the \textit{User Personalised Layers} are only stored locally and never shared with the server.

\subsection{Semi-supervised learning}
\label{subsec:class_and_lab}
In the following, we describe how each client semi-automatically provide labels to the stream of unlabeled sensor data. \acronym~ relies on a combination of two semi-supervised learning techniques: \textit{Active Learning} and \textit{Label Propagation}.

%On the one hand \textit{Active learning} is described in section \ref{subsec:active_learning} and involves triggering questions to the users in order to obtain the labels of those samples classified with low confidence. On the other hand the \textit{label propagation} algorithm \ref{subsec:label_spreading} allows spreading the collected labels to the unlabelled data aiming to train locally the model with an augmented number of samples.

\subsubsection{Active Learning}
\label{subsec:active_learning}
%In our work, we aim to design a system much more suitable for everyday life by dramatically reducing the users' effort to collect labelled examples.
An active learning process requires the user feedback about her currently performed activity when there is uncertainty in the classifier's prediction. 
The intuition is the following: unlabeled data samples for which the classification confidence is significantly low would have the most impact in improving the classifier if the label were available (i.e., they are the more informative ones).

\acronym\ relies on a state-of-the-art non-parametric active learning approach called \textit{VAR-UNCERTAINTY}~\cite{vzliobaite2013active}.
This approach compares the prediction confidence with a threshold $\theta\in[0,1]$ that is dynamically adjusted over time. Initially, $\theta$ is initialised to $\theta=1$.
Let $\mathbf{A} = \{A_1, A_2, \dots, A_n\}$ be the set of target activities. Given the probability distribution over the possible activities of the current prediction $\langle p_1, p_2, \dots, p_n \rangle$, we denote with $p^\star = \max_{i} p_i$ the probability value of the most likely activity $A^\star \in \mathbf{A}$ (i.e., the predicted activity).
If $p^\star$ is below $\theta$, we consider the system uncertain about the current activity performed by the user. In this case, an active learning process is started by asking the user the ground truth $A^f \in \mathbf{A}$ about the current activity. The feedback $A^f$ is stored in the \textit{Feature Vectors Storage}. When $A^f = A^\star$, it means that the most likely activity $A^\star$ is actually the one performed by the user, and hence the threshold $\theta$ is decreased to reduce the number of questions. On the other hand, when $A^f \neq A^\star$, $\theta$ is increased. More details about this active learning strategy can be found in~\cite{vzliobaite2013active}.
The pseudo-code of classification and active learning is described in Algorithm~\ref{alg:clientactive}

\begin{figure}[h!]
	\centering
	\includegraphics[scale=0.12]{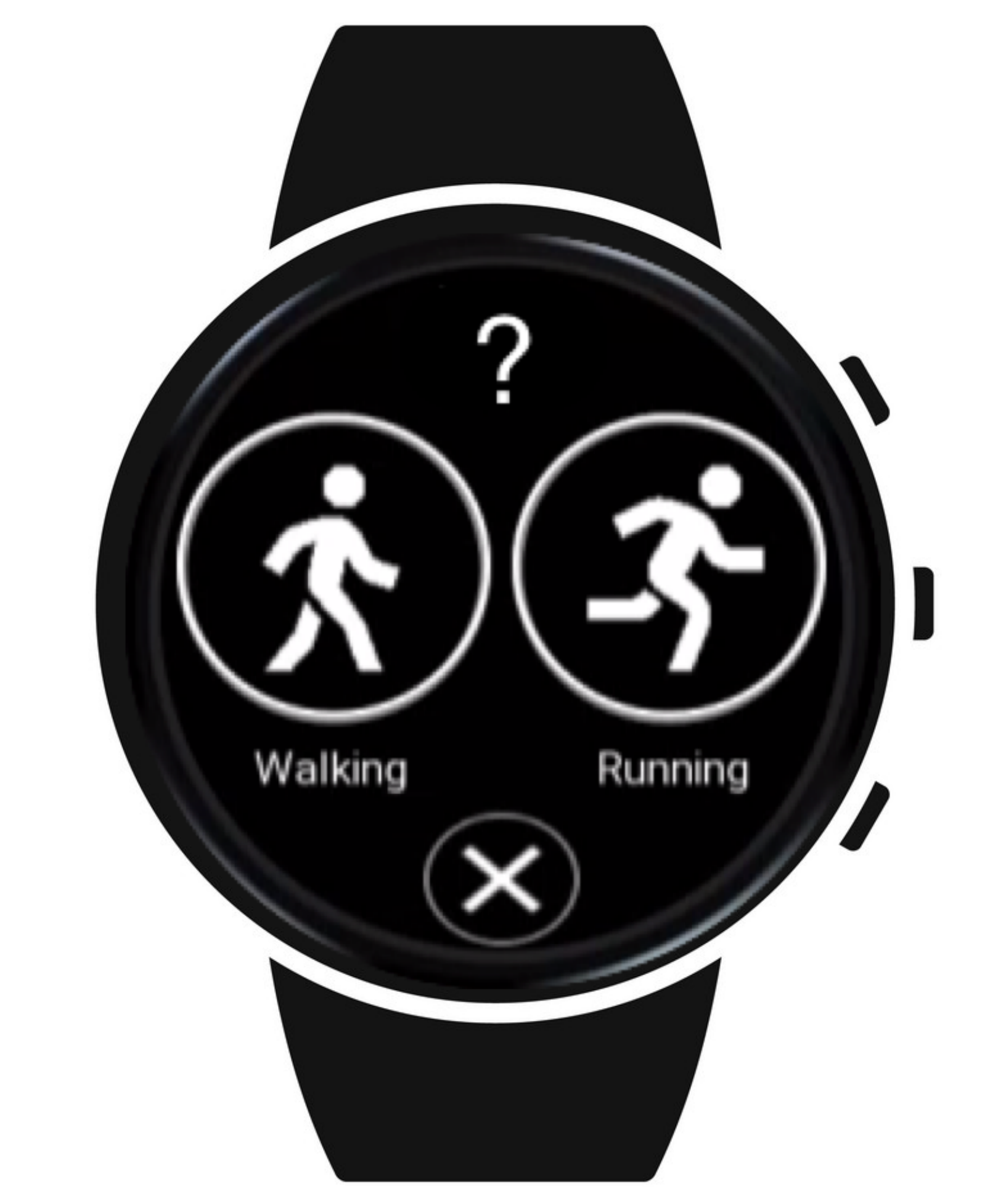}
	\caption{Example of an active learning interface for smart-watches.}
	\label{fig:al_interface}
\end{figure}

We assume that active learning queries are prompted to the user in real-time through a dedicated application, thanks to a user-friendly interface. Each query asks the user to choose the activity that she is currently performing among the possible ones.
For the sake of usability, \acronym\ only presents a couple of alternatives taken from the most probable activities. Figure~\ref{fig:al_interface} shows a screenshot of an active learning application that we implemented for smart-watches in another research work.

\begin{algorithm}
\caption{Client side - Classification and data labeling}\label{alg:clientactive}
%\hspace*{\algorithmicindent} \textbf{Input} \\
%\hspace*{\algorithmicindent} \textbf{Output} 
\begin{algorithmic}[1]
\State $sm \Leftarrow$ Shareable Model 
\State $pm \Leftarrow$ Personalized Model

\State receive pre-trained $w^G$ from the server
\State $sm \Leftarrow w^G$
\State $pm \Leftarrow w^G$

\For{each feature vector $fv$ computed in real-time from sensor data}

  \State $\vec{p} \Leftarrow$ probability distribution over the activities predicted by $pm$ on $fv$
  \State output the most likely activity according to $\vec{p}$
  \If{a feedback is needed according to VAR-UNCERTAINTY~\cite{vzliobaite2013active}}
  \State $l \Leftarrow$ activity label from the user
  \State add $(fv,l)$ to the Feature Vectors Storage
  \Else
  \State add $(fv,-)$ to the Feature Vectors Storage \Comment{unlabeled data point}
  \EndIf

\EndFor
\end{algorithmic}
\end{algorithm}

\subsubsection{Label Propagation}
\label{subsec:label_spreading}
The major drawback of active learning is that the queries may interrupt the user while performing an activity. 
%during her daily life.
%Hence, for the sake of user experience, the number of those queries should be low.
In order to reduce the interaction with the user and, at the same time, to train the local models with a larger amount of labeled data, \acronym~ also relies on label propagation. 
%with the objective of improving the recognition rate and, at the same time, to further reduce the number of active learning questions. 
The Label Propagation process is started when the server requires to update the global model (see Algorithm~\ref{alg:serverside}).
Given a set of labeled and unlabeled data points, the goal of label propagation is to automatically spread labels to a portion of unlabeled data~\cite{zhou2004learning}. The intuition behind label propagation is that data points close in the feature space likely correspond to the same class label. The Label Propagation model of \acronym\ is a fully connected graph $g = (V,E)$ where the nodes $V$ are all the data samples in the \textit{Feature Vectors Storage} and the weight on each edge in $E$ is the similarity between the connected data points.
%is to construct from the features vector matrix of all samples (labelled and not) a fully connected graph $g = (V,E)$ where nodes $V = {1,…,n}$ represent the given sample and edges $E$ are weighted with the similarity between the connected nodes.
In the literature, this similarity is usually computed using K-Nearest Neighbors (KNN) or Radial Basis Function Kernel (RBF kernel). \acronym\ relies on the RBF kernel due to its trade-off between computational costs and accuracy~\cite{widmann2017graph}. Formally, the RBF kernel function is defined as $K(x, x') = e ^{- \gamma||x - x' ||^2}$ where $||x - x' ||^2$ is the squared Euclidean distance between the feature vectors of two nodes $x$ and $x'$ (where $x'$ is a labeled node), and $\gamma \in \mathbb{R}+$. Hence, the value of the RBF kernel function increases as the distance between data points decreases. The kernel is used to perform inductive inference to predict the labels on unlabeled data points, based on a threshold on the similarity between the nodes. This process is repeated until convergence (i.e., when there are no more unlabeled data points for which label propagation is reliable based on the threshold).
%which will produce a fully connected graph with a dense matrix.

%In \acronym, the Label Propagation model is initialized during the pre-training phase, as described in Section~\ref{subsec:pre_train}, and it is distributed to all participating devices.
%instance of a label propagation algorithm using a small fraction of the labelled samples collected in a publicly available activity recognition dataset (that we previously called pre-training set).
In \acronym, the Label Propagation model (i.e., the graph) is initialized with the labeled data points of the \textit{pre-training dataset}. Moreover, this model is personal and never shared with other users nor with the server.
%(labelled thanks to active learning or not). 
%Thus, when the device participates to a communication round, %the label propagation algorithm can be executed locally, propagating the labels to the unlabelled data.
%the newly labelled data acquired are used the train the local instance of the global model as we previously described.

%% file: sections/experiments.tex
In this section, we describe in detail the extensive experimental evaluation that we carried out to quantitatively assess the effectiveness of \acronym. First, we describe the public datasets that we considered in our experiments. Then, we present our novel evaluation methodology that aims at assessing both the generalization and personalization capabilities of \acronym. Finally, we  discuss the results that we obtained 
%by applying our evaluation methodology 
on the target datasets.

\subsection{Datasets}

Since FL makes sense when many users participate in collaboratively training the global model, we considered publicly available datasets of physical activities (performed both in outdoor and indoor environments) that were collected involving a significant number of subjects. %In particular, our target datasets contain labeled inertial sensor data from personal mobile devices.
However, there are only a few public datasets with these characteristics. One of them is MobiAct~\cite{vavoulas2016mobiact}, which includes labeled data from $60$ different subjects with high variance in age and physical characteristics. The dataset contains data from inertial sensors (i.e., accelerometer, gyroscope, and magnetometer) of a smartphone positioned in a trousers' pocket freely chosen by the subject in any random orientation. $73\%$ of the participants were male, while $27$ are female. The subjects' age spanned between $20$ and $47$  (average: $26$), the height ranged from $160$cm to $189$cm (average: $175$), and the weight varied from $50$kg to $120$kg (average: $76$). The adopted data acquisition frequency is the highest enabled by the sensors of the selected smartphone (i.e., at most 200Hz). Due to its characteristics, this dataset was also used in other works that proposed FL applied to HAR (e.g., the work presented in~\cite{wu2020fedhome}).
In our experiments, we considered the following physical activities~\footnote{Note that we omitted from MobiAct those physical activities with a limited number of samples. Indeed, as we will explain later, our evaluation methodology requires to partition the data of each user. Activities with a small number of samples would be insufficiently represented in each partition and hence they are not suitable for our evaluation. We believe that this problem is only related to this specific dataset and that, in realistic settings, even short activities would be represented by a sufficient number of samples.}: \textit{standing}, \textit{walking}, \textit{jogging}, \textit{jumping}, and \textit{sitting}. The distribution of activity labels in Mobiact is illustrated in Table~\ref{tab:distribution_mob}.

We also consider the well-known WISDM dataset~\cite{kwapisz2011activity}. This dataset has been widely adopted as benchmark for HAR. WISDM contains accelerometer data (sampling rate 20HZ) collected from a smartphone located in the front pants leg pocket of each subject during activity execution. WISDM includes data from $36$ subjects. The data collection was supervised by one of the WISDM team members to ensure the quality of the collected data. The activities included in this dataset are the following: \textit{walking}, \textit{jogging}, \textit{sitting}, \textit{standing}, and \textit{taking stairs}. The distribution of activity labels in WISDM is illustrated in Table~\ref{tab:distribution_wisdm}. Unfortunately, further information about the participants like gender, age, and weight distribution is not publicly available.

\begin{table}[h!]
  \begin{minipage}{.5\linewidth}
    \centering
    \begin{tabular}{lc}
\hline
\textbf{Activity} & \begin{tabular}[c]{@{}c@{}}percentage\\ of samples\end{tabular} \\ \hline
Standing              & 44\%                                                            \\
Walking              & 42\%                                                            \\
Jogging            & 6\%                                                             \\
Jumping             & 6\%                                                             \\
Sitting              & 2\%                                                             \\ \hline
                 & \multicolumn{1}{l}{}                                            \\
TOTAL            & \multicolumn{1}{l}{18.654 samples}                              \\ \hline
\end{tabular}
    \caption{MobiAct: distribution of\\the considered activities}\label{tab:distribution_mob}
  \end{minipage}
  \begin{minipage}{.5\linewidth}
    \centering
    \begin{tabular}{lc}
\hline
\textbf{Activity} & \begin{tabular}[c]{@{}c@{}}percentage\\ of samples\end{tabular} \\ \hline
Walking        & 38\%                                                            \\
Jogging        & 30\%                                                            \\
Sitting        & 6\%                                                             \\
Standing       & 5\%                                                             \\
Upstairs       & 11\%                                                            \\
Downstairs     & 10\%                                                             \\ \hline
               & \multicolumn{1}{l}{}                                            \\
TOTAL          & \multicolumn{1}{l}{13.726 samples}                              \\ \hline
\end{tabular}
    \caption{WISDM: distribution of\\the considered activities}\label{tab:distribution_wisdm}
  \end{minipage}%
\end{table}

\subsection{Evaluation methodology}
\label{subsec:evaluationmethodology}

In the following, we describe 
%our novel 
the methodology that we 
%accurately 
designed to evaluate the effectiveness of \acronym, both in terms of personalization and generalization.
We split each dataset into three partitions that we call $Pt$, $Tr$, and $Ts$.
%assigning to each partition the data collected from respectively $pt, ts$ and $tr$ different users. 
The partition $Pt$ (i.e., pre-training data) contains data of users that we only use to initialize the global model. 
$Tr$ (i.e., training data) is the dataset partition that includes data of users who participate in FL. Finally, $Ts$ (i.e., test data) is a dataset partition that includes data of left-out users %that do not contribute to the global model, and %
that we only consider to periodically evaluate the generalization capabilities of the global model. In our experiments, we randomly partition the users as follows: $15\%$ whose data will populate $Pt$, $65\%$ whose data will populate $Tr$, and $20\%$ whose data will populate $Ts$.

%In order to assess how the global model evolves over different FL iterations, 
We partition the data for each user in $Tr$ into $sh$ shards of equal size. 
%Each shard is a batch of data that is collected from a user during a specific time interval. 
In realistic scenarios, each shard should contain data collected during a relatively long time period (e.g., a day) where a user executes many different activities. However, the considered datasets only have a limited amount of data for each user (usually less than one hour of activities for each user). 
Hence, we generate shards as follows. Given a user $u \in Tr$, we randomly assign to each shard a fraction $\frac{1}{sh}$ of the available data samples associated to $u$ in the dataset. Note that each data sample of a user is associated to exactly one shard. This mechanism allows us to simulate the realistic scenario described before, where users perform several types of activities in each shard. \\

\paragraph{Evaluation Algorithm}
In the following, we describe our novel evaluation methodology step by step.
First, the labeled data in $Pt$ are used to initialize the global model, that is then distributed to the devices of all the users in $Tr$ that will use it as the first version of the \textit{Personalized Model}.
%2
We evaluate the recognition capabilities of the initial pre-trained global model on the partition $Ts$ in terms of F1 score. This assessment allows us to measure how the initial global model generalizes on unseen users before any FL step.
%improvements over those users that never take part to train the model.

As we previously mentioned, for each user, we partition its data samples in $Tr$ into exactly $sh$ shards.
For the sake of evaluation, we assume a synchronous system in which the shards of the different users in $Tr$ are actually temporally aligned and  occur simultaneously (i.e, the first shards of every user occur at the same time interval, the second shards of every user occur at the same time interval, and so on).  Note that, in the considered datasets, each user has a different data distribution and a different number of samples. Hence, within a specific shard, each client contributes with data collected considering its personal distribution.
%Hence, during the parallel execution of a shard 
The evaluation process is composed by $sh$ iterations, one for each shard.
Considering the $i$-th shard we proceed as follows:
\begin{enumerate}
    \item The devices of the users in $Tr$ exploits the \textit{Personalized Model} to classify the continuous stream of inertial sensor data in its shard. We use the classification output to evaluate the recognition rate in terms of F1 score providing an assessment of personalization. Note that, during this phase, we also apply our active learning strategy and we keep track of the number of triggered questions.
    \item When all data in the shard have been processed (by all devices), the server starts a number $r$ of communication rounds with a subset of the devices in order to update the global weights. Each round is implemented as follows:
    \begin{enumerate}
        \item The server randomly selects a certain percentage $p\%$ of users in $Tr$ and sends to their devices the last update of the global weights.
        \item Each user's device, by receiving the global weights, applies Label Propagation (See Section~\ref{subsec:label_spreading}) and uses the newly labeled data to train its \textit{Shareable Model}. After training, the resulting weights are transmitted to the server.
        \item The server merges the received weights obtaining a new version the global model weights.
        \item We evaluate in terms of F1 score the recognition rate of the resulting global model on the left-out users in $Ts$ (providing an assessment of generalization).
    \end{enumerate}
    \item After the execution of all the communication rounds, each users' device:
    \begin{enumerate}
          \item replaces the weights of the \textit{Shareable Model} and \textit{Personalized Model} with the ones of the latest global model
          \item fine-tunes the \textit{Personalized Model} with labeled data from active learning and label propagation
          \item starts the personalization process described in Section~\ref{sec:personalization}.
    \end{enumerate}
\end{enumerate}

%The above mentioned steps are repeated for each parallel execution of a shard.
Note that our evaluation methodology introduces several levels of randomness: assigning users to $Ts$, $Tr$ and $Pt$; assigning data samples to shards; selecting devices at each communication round. We iterate experiments $10$ times and average the results in order to make our estimates more robust.

\subsection{Results}

In the following, we report the results of the evaluation of \acronym. 

\subsubsection{Classification model and hyper-parameters}

As explained and motivated in Section~\ref{subsec:model}, our classification model is a fully-connected deep neural network. The network consists of four fully connected layers having respectively $128$, $64$, $32$, and $16$ neurons, and a softmax layer for classification. We use Adam~\cite{kingma2017adam} as optimizer. The choice of this specific network architecture is due to the good performance reported in the federated HAR literature~\cite{wu2020fedhome}.
%: 1) our method does not rely on feature learning that is not suitable in our semi-supervised setting
%, 2) training a fully-connected network is more suitable for mobile devices since it requires less computational resources with respect to more complex networks (e.g., CNN, LSTM), 
%and 3) 
As hyper-parameters, we empirically chose $w=4s$, $p=30\%$, $r=10$, $l=2$,  $sh=3$, and  $10$ local training epochs with a batch size of $30$ samples. These hyper-parameters have been empirically determined based on data in $Ts$. 
The low number of epochs and communication rounds is due to the small size of the public datasets. This also limits the data in each shard.  In a large scale deployment, these parameters should be accurately calibrated.

\subsubsection{Impact of semi-supervised learning}

\begin{figure}[h!]
\centering
    \subfloat[Average F1 score]{{\includegraphics[width=5cm]{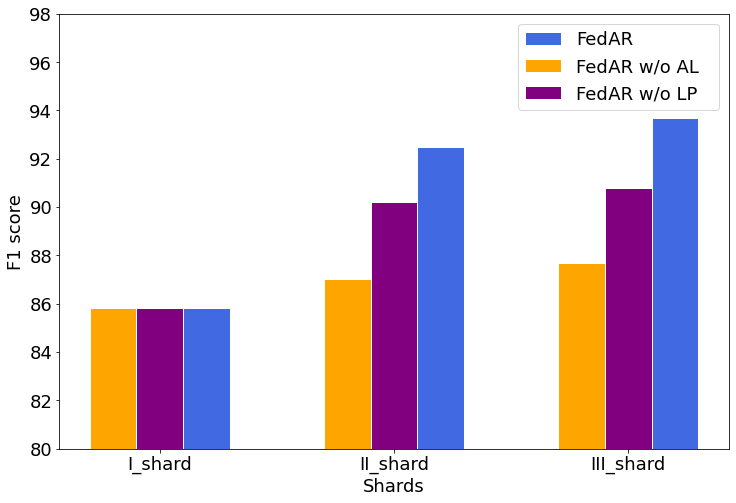} 
	\label{fig:mobiact_semisupervised_train_f1}
	}}
	\qquad
		\subfloat[Average percentage of active learning questions.]{{\includegraphics[width=5cm]{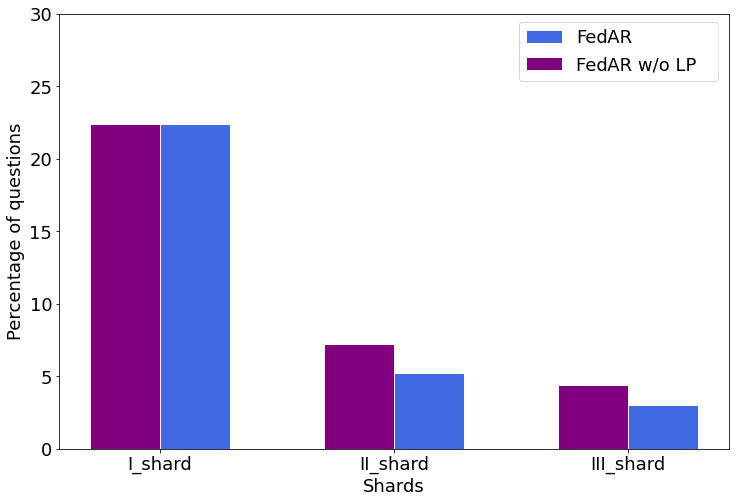} 
	\label{fig:mobiact_semisupervised_train_q}
	}
	}
	\caption{MobiAct: The impact of label propagation and active learning on the subjects that participated in the FL process.}
	\label{fig:mobiact_semisupervised_train}
\end{figure}

\begin{figure}[h!]
\centering
    \subfloat[Average F1 score]{{\includegraphics[width=5cm]{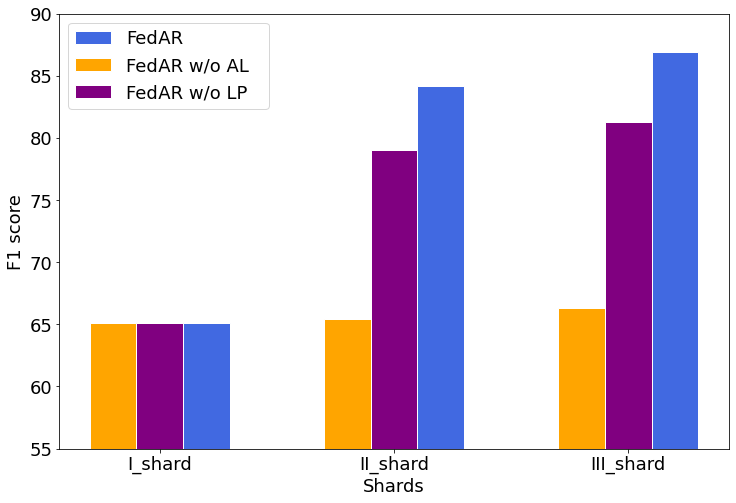} 
	\label{fig:wisdm_semisupervised_train_f1}
	}}
	\qquad
		\subfloat[Average percentage of active learning questions.]{{\includegraphics[width=5cm]{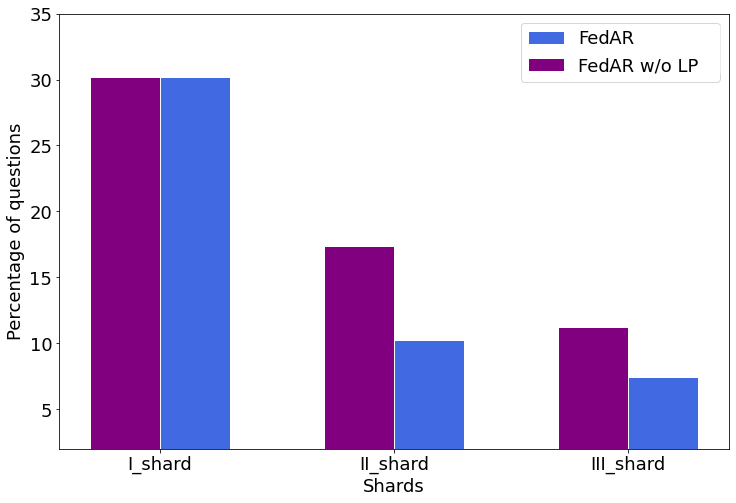} 
	\label{fig:wisdm_semisupervised_train_q}
	}
	}
	\caption{WISDM: The impact of label propagation (LP) and active learning (AL) on the subjects that participated in the FL process.}
	\label{fig:wisdm_semisupervised_train}
\end{figure}

Figure~\ref{fig:mobiact_semisupervised_train} and Figure~\ref{fig:wisdm_semisupervised_train} show how the F1 score and the percentage of active learning questions change at each shard for the users in $Tr$.

We observe that the F1 score significantly improves shard after shard, while the number of active learning questions decreases. 
Averaging the results of both datasets, the number of active learning questions at the first shard is around $25\%$, while at the last shard is only around $5\%$. This result indicates that our method continuously improves the recognition rate with a limited amount of labels provided by the users. Moreover, the continuous decrease of the number of questions militates for the usability of our method, which will prompt fewer and fewer questions with time.
These figures also show the impact of combining active learning with label propagation. Without label propagation, active learning alone leads to a lower recognition rate and a higher number of questions. This means that the labeled data points derived by label propagation positively improve the activity model.
On the other hand, we observe that label propagation leads to unsatisfying results without active learning. Indeed, the labeled samples obtained by active learning represent informative data that are crucial for label propagation. 
Hence, the evaluation on both datasets confirms that the combination of active learning and label propagation leads to the best results.

\begin{figure}[h!]
	\centering
	\includegraphics[scale=0.28]{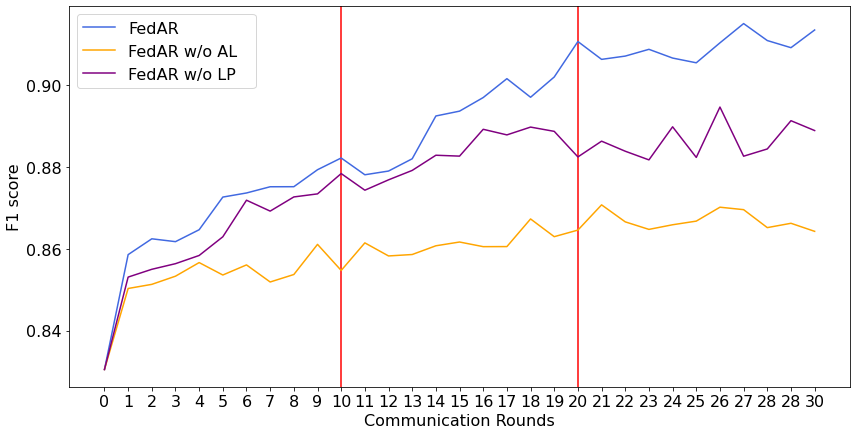}
	\caption{MobiAct: the trend of F1 score on the left-out users after each communication round. This Figure also shows the impact of active learning and label propagation. Each red line marks the end of a shard.}
	\label{fig:mobiact_semisupervised_test}
\end{figure}

\begin{figure}[h!]
	\centering
	\includegraphics[scale=0.28]{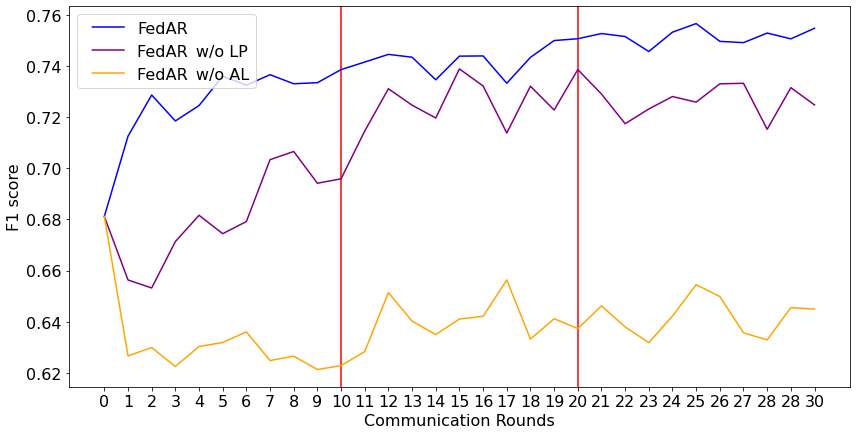}
	\caption{WISDM: the trend of F1 score on the left-out users after each communication round. This Figure also shows the impact of active learning and label propagation. Each red line marks the end of a shard.}
	\label{fig:wisdm_semisupervised_test}
\end{figure}

In Figure~\ref{fig:mobiact_semisupervised_test} and Figure~\ref{fig:wisdm_semisupervised_test} we show the generalization capability of the global model on left-out users (i.e., users in partition $Ts$) after each communication round performed during the FL process with the users in $Tr$. 

The red lines mark the end of each shard. The results indicate that the federated model constantly improves also for those users that did not contribute with training data, even if the active learning questions continuously decrease. 
These plots also confirm that the combination of label propagation and active learning leads to the best results on both datasets.

\subsubsection{\acronym\ versus approaches based on fully labeled data}

We compared our approach with two existing FL methods based on fully labeled data.
The first one is the well-known FedAVG~\cite{mcmahan2017communication}, which is the most common FL method in the literature. FedAVG simply averages the model parameters derived by the local training on the participating nodes (without any personalization).

The second method that we use for comparison is called FedHealth~\cite{chen2020fedhealth}. This is one of the first FL approaches proposed for activity recognition on wearable sensors data. Similarly to our approach, FedHealth applies personalization using transfer learning.

Since~\acronym\ considers a limited amount of available labeled data, our goal is to achieve a recognition rate that is as close as possible to the one obtained by solutions that assume full availability of annotations.

For the sake of fairness, in our experiments we adapted FedAVG and FedHealth to use the same neural network that we use in \acronym. Hence, we performed our experiments using our evaluation methodology by simulating that, for FedAVG and FedHealth, each node has the ground truth for each data sample on each shard. Hence, the evaluation of those methods does not include active learning and label propagation. Moreover, differently from \acronym, FedAVG and FedHealth only use a single local model.

\begin{figure}[t]
\centering
    \subfloat[MobiAct]{{\includegraphics[width=7cm]{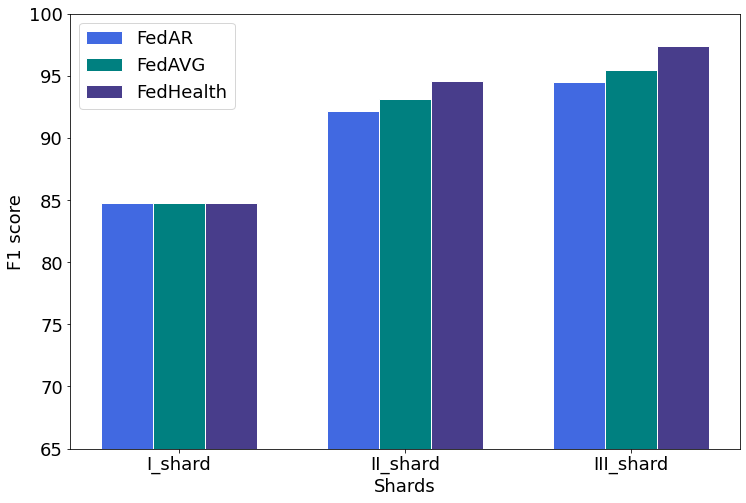} 
	\label{fig:mobiact_baselines}
	}}
	\qquad
		\subfloat[WISDM]{{\includegraphics[width=7cm]{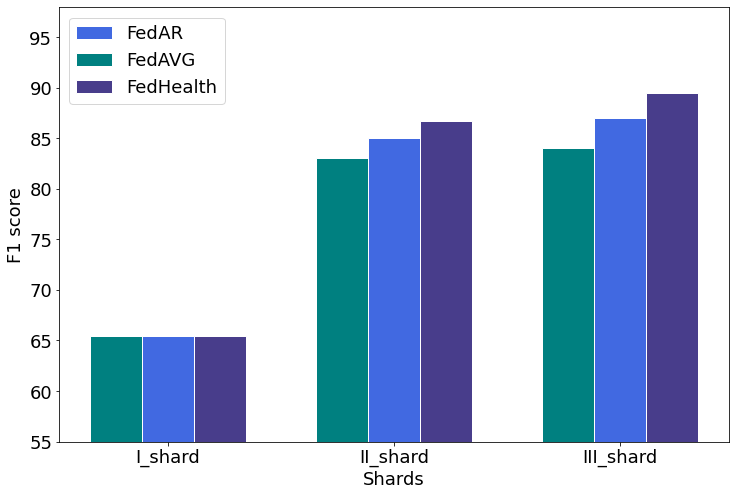} 
	\label{fig:wisdm_baselines}
	}
	}
	\caption{Comparison of \acronym\ with methods based on fully labeled data.}
	\label{fig:baselines}
\end{figure}

The results of this comparison for the users in $Tr$ (i.e., the ones that actively participated in the FL process) are reported in Figure~\ref{fig:mobiact_baselines} and Figure~\ref{fig:wisdm_baselines}.

From these plots, we observe that \acronym\ converges to recognition rates that are similar to solutions based on fully labeled data at each shard. The advantage of \acronym\ is that it can be used for realistic HAR deployments where the availability of labeled data is scarce. Despite a reduced number of required annotations, \acronym\ performs even better than FedAVG on the WISDM dataset, while on MobiAct it performs slightly worse. Moreover, \acronym\ is only $\approx 3\%$ behind FedHealth on both datasets.

\subsubsection{\acronym\ performance on each activity}

\begin{figure}[h!]
\centering
    \subfloat[MobiAct]{{\includegraphics[width=7cm]{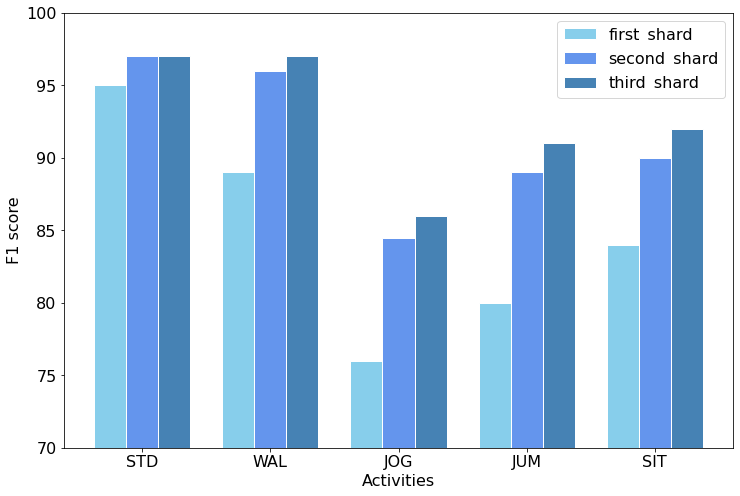} 
	\label{fig:mobiact_activity_details}
	}}
	\qquad
		\subfloat[WISDM]{{\includegraphics[width=7cm]{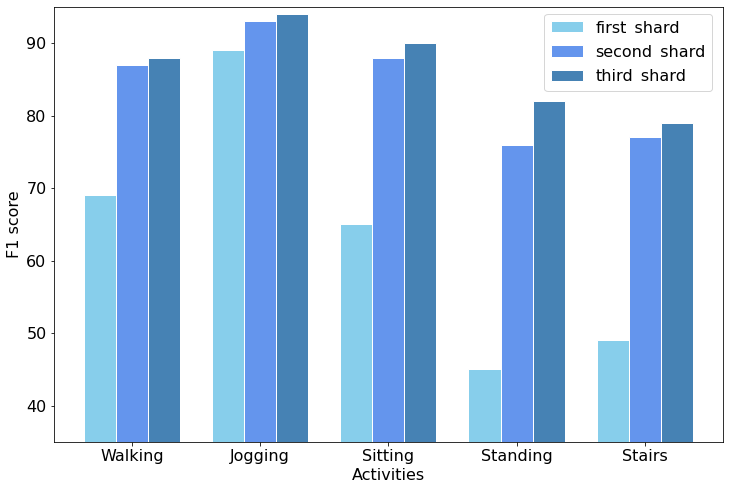} 
	\label{fig:wisdm_activity_details}
	}
	}
	\caption{F1 score at each shard for each activity on the users that participated in the FL process.}
	\label{fig:activity_details}
\end{figure}

Figure~\ref{fig:activity_details} shows how the recognition rate improves between shards for each activity for the users in $Tr$ on both datasets.

We observed an improvement of the recognition rate shard after shard for each considered activity. The only exception is the \textit{standing} activity on the MobiAct dataset in the third shard, which maintains the same F1 score. 

In general, the greatest improvement occurs between the first and the second shards. This is due to the fact that, in the first shard, activities are recognized using the initial global model only trained with the \textit{pre-training dataset}. Starting from the second shard, classification is performed with the \textit{Personalized Model} updated thanks to FL and personalized using our transfer learning approach.

\subsubsection{Impact of personalization}

\begin{figure}[h!]
\centering
    \subfloat[Average F1 score]{{\includegraphics[width=7cm]{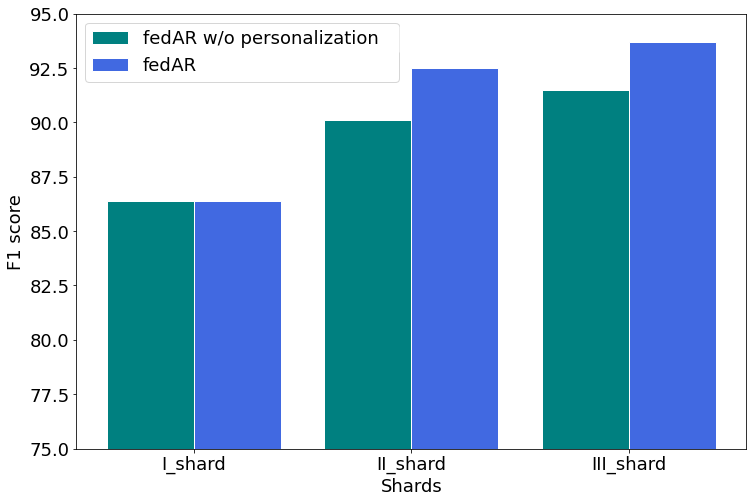} 
	\label{fig:mobiact_train_f1}
	}}
	\qquad
		\subfloat[Average percentage of active learning questions]{{\includegraphics[width=7cm]{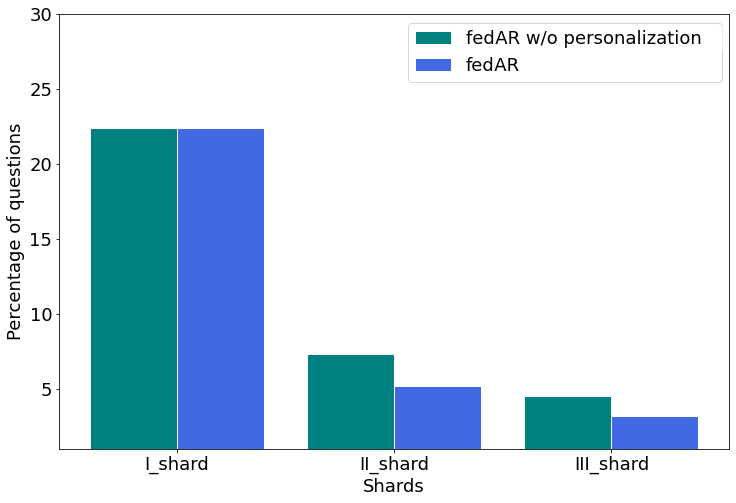} 
	\label{fig:mobiact_train_q}
	}
	}
	\caption{MobiAct: results on the users that participated to the FL process for each shard, with and without personalization.}
	\label{fig:mobiact_personalization}
\end{figure}

\begin{figure}[h!]
\centering
    \subfloat[Average F1 score]{{\includegraphics[width=7cm]{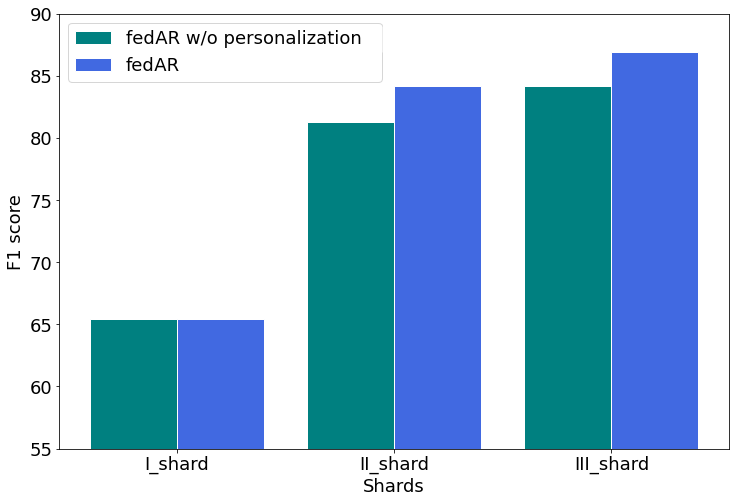} 
	\label{fig:wisdm_train_f1}
	}}
	\qquad
		\subfloat[Average percentage of active learning questions]{{\includegraphics[width=7cm]{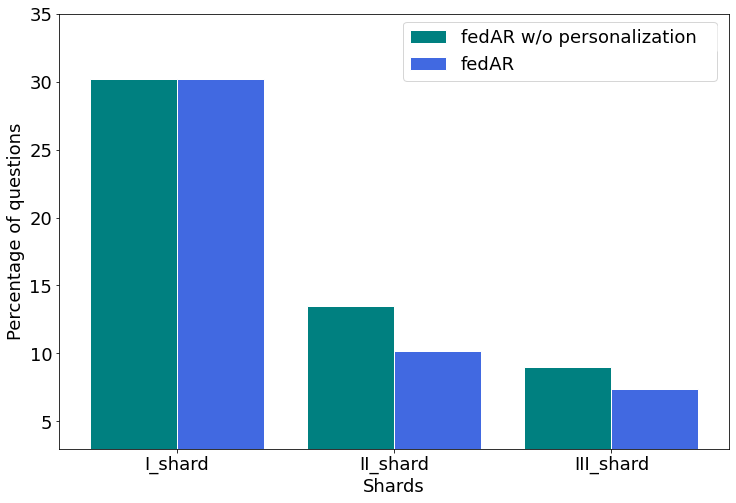} 
	\label{fig:wisdm_train_q}
	}
	}
	\caption{WISDM: results on the users that participated to the FL process for each shard, with and without personalization.}
	\label{fig:wisdm_personalization}
\end{figure}

Figure~\ref{fig:mobiact_personalization} and Figure~\ref{fig:wisdm_personalization} show the impact of the \acronym\ personalization strategy based on transfer learning. This evaluation is performed on the users in the $Tr$ partition.

As expected, fine-tuning the personal models leads to an improvement both on the recognition rate and on the number of questions in active learning. Note that, during the first shard, classification is performed using the weights derived from the \textit{pre-trained dataset} and personalization is applied starting from the second shard.

\subsubsection{Fully connected vs Convolutional models}
\label{subsubsec:mlp_vs_cnn}

The classification model in \acronym\ is a fully connected network (we will refer it as MLP\footnote{MultiLayer Perceptron} for the sake of brevity) that receives as input handcrafted feature vectors.
Nonetheless, Convolutional Neural Networks (CNNs) proved to be very effective in fully supervised HAR approaches, since they can automatically learn features from raw data~\cite{cruciani2019comparing}. 

\begin{figure}[h!]
\centering
    \subfloat[WISDM]{{\includegraphics[width=3cm]{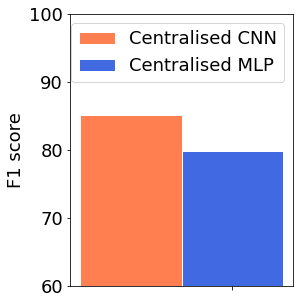} 
	\label{fig:w_centralised_CNN_}
	}}
	\qquad
		\subfloat[MobiAct]{{\includegraphics[width=3cm]{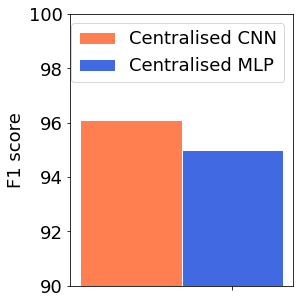} 
	\label{fig:M_centralised_CNN_}
	}
	}
	\caption{Centralized setting: MLP vs CNN based on leave-one-subject-out cross-validation.}
	\label{fig:centralised_CNN_FCN}
\end{figure}

We performed a preliminary experiment to compare MLP and CNN in a fully supervised centralized approach using a leave-one-subject-out cross-validation.
As CNN architecture, we consider the one recently proposed in~\cite{wan2020deep} since it proved to be one of the most effective for HAR. Figure~\ref{fig:centralised_CNN_FCN} shows the outcome of this comparison. We observe that, considering a fully-supervised centralized setting, CNN is more effective on both datasets.

%In the literature, both convolutional network (CNN)  and fully connected network (MLP) \cite{wu2020fedhome} shown to be effective in the field of HAR. Considering the features used to train the recognition model, the convolutional layers of the CNN enable automatically extracting them from the raw sensors data. Differently, the fully connected layers of the MLP require handcrafted features.

 %leave-one-subject-out cross-validation approach to evaluate and compare the F1 score obtained by using CNN and MLP on the considered datasets. In particular, we adopted the same CNN configuration proposed by F. Cruciani in  as it exhibited good performances for HAR.  
 %Regarding the MLP network, we leverage only four fully connected layers having respectively 128, 64, 32, and 16 neurons, and a softmax layer for classification. 
%The histograms reported in Figure \ref{fig:centralised_CNN_FCN} show that in the considered fully-supervised centralised setting, the CNN network results more effective than MLP in terms of recognition rate both on MobiAct and (especially) on WISDM.

\begin{figure}[h!]
\centering
    \subfloat[Average F1 score]{{\includegraphics[width=5cm]{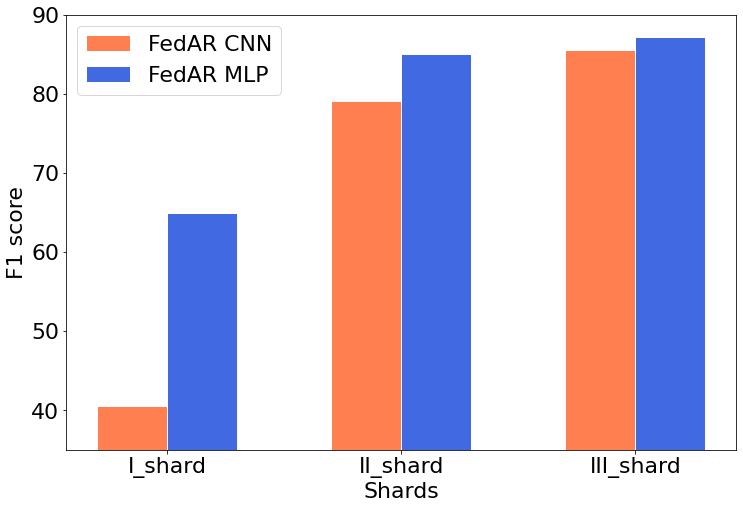} 
	\label{fig:W_train_F1_FCN_CNN}
	}}
	\qquad
		\subfloat[Average percentage of active learning questions]{{\includegraphics[width=5cm]{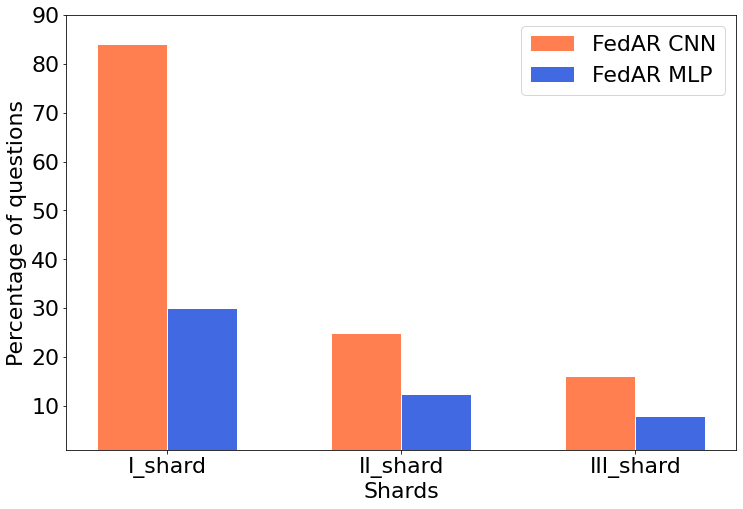} 
	\label{fig:W_train_Q_FCN_CNN}
	}
	}
	\caption{WISDM: results on the users that participated to the FL process for
each shard using both CNN and MLP networks}
	\label{fig:W_train_CNN_FCN}
\end{figure}

\begin{figure}[h!]
\centering
    \subfloat[Average F1 score]{{\includegraphics[width=5cm]{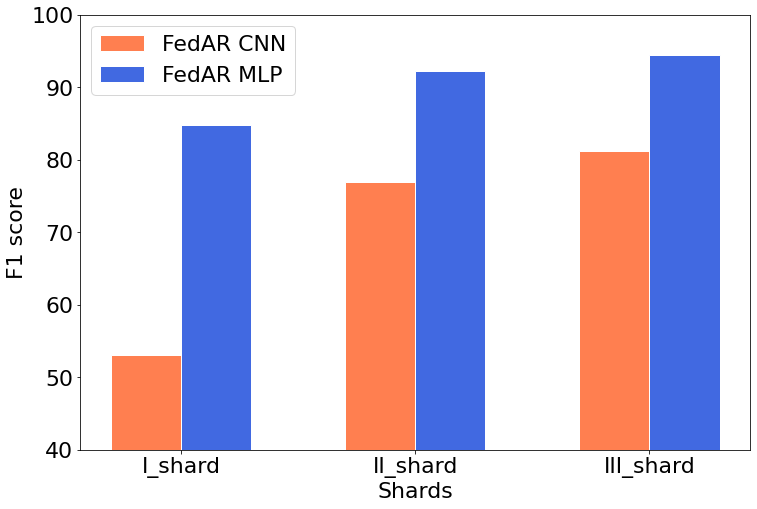} 
	\label{fig:M_train_F1_FCN_CNN}
	}}
	\qquad
		\subfloat[Average percentage of active learning questions]{{\includegraphics[width=5cm]{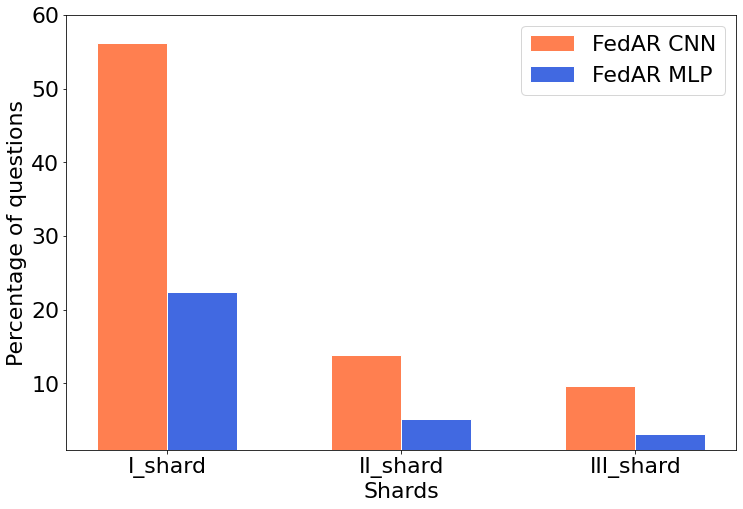} 
	\label{fig:M_train_Q_FCN_CNN}
	}
	}
	\caption{MobiAct: results on the users that participated to the FL process for
each shard using both CNN and MLP networks}
	\label{fig:M_train_CNN_FCN}
\end{figure}

However, we observed that CNN struggles in learning reliable features considering our federated and semi-supervised setting, since the amount of labeled data to train the classifier is limited (cold start issue). Figures~\ref{fig:W_train_CNN_FCN} and~\ref{fig:M_train_CNN_FCN} show the comparison of \acronym\ using our MLP model with handcrafted features and the CNN model. On both datasets, MLP quickly reaches a higher F1 score with respect to CNN with a significantly lower number of active learning queries. Since features are computed a priori, the MLP model can immediately focus on training the classification layers rather than learning features.
Hence, these results motivate our choice of adopting a MLP model with handcrafted features in~\acronym.

%than CNN both in terms of the F1 score and the percentage of questions to the users. In particular, it is clear that (especially in the initial shards) the MLP setup quickly reaches a higher F1 score with respect to the CNN setup, also triggering a very limited number of questions. Thus, as we expected, the CNN network suffers from the cold start issue, and hence requires triggering a large number of questions to the users in order to obtain enough annotated samples to train both the convolutional and the classification layers. Differently, considering the MLP setup, the features are computed a priory, and hence the model can start right away to learn the parameters of the classification layers instead of spending resources to extract features. 

%A comparable behavior is illustrated by the histograms in figures \ref{fig:test_CNN_FCN} which represent the recognition rate obtained on the left-out users using CNN and MLP.  

%Thus, considering our experimental evaluation, we decided to use in \acronym\ a simple MLP that with no layer specialised in features learning.

%% file: sections/discussion.tex
\subsection{Generality of the approach}
\label{subsec:generality}

While we designed \acronym~ with wearable-based activity recognition as target application, we believe that this combination of semi-supervised and FL can be applied also to many other applications. Our method is suitable for human-centered classification tasks that include the following characteristics: 

\begin{itemize}
    \item There is a large number of clients that participate in the FL process.
    \item Classification needs to be performed on a continuous data stream, where labels are not naturally available.
    \item Each node generates a significant amount of unlabeled data.
    \item It is possible to periodically obtain the ground truth by delivering active learning questions to users that are available to provide a small number of labels. Note that, in real-time applications like HAR, for the sake of usability active learning questions should be prompted temporally close to the prediction.
    \item It is possible to obtain a limited training set to initialize the global model. Hence, a small group of volunteers should be available (in an initial phase) for annotated data acquisition.
    \item The nodes should be capable of computing training operations. Clearly, nodes can also rely on trusted edge gateways/servers (like proposed in~\cite{wu2020fedhome}).
\end{itemize}

%\subsection{Initialization of the global model}

%In our experimental evaluation, we use a portion of the dataset to initialize the recognition model. However, in real-world settings, the global model should be initialized using publicly available HAR datasets. 
%However, it is not clear who should perform the initial training of the model. Indeed, the initial training phase would require the knowledge of the structure of the classifier, while the cloud server should only be aware of the model weights for the sake of privacy. We hypothesize that an external service could be used to define the structure of the neural network and initialize it with the pre-training set. Then, the resulting weights are provided to the cloud server.

\subsection{Privacy concerns}

Despite FL is a significant step towards protecting user privacy in distributed machine learning, the shared model weights may still reveal some sensitive information about the participating users~\cite{nasr2018comprehensive,shokri2017membership}.
Similarly to other works, \acronym\ uses Secure Multiparty Computation (SMC)~\cite{bhowmick2018protection,cramer2000general} to mitigate this problem. 
However, other approaches have been proposed, including differential privacy (DP)~\cite{agarwal2018cpsgd,dwork2008differential}, and hybrid approaches that combine SMC and DP~\cite{truex2019hybrid}.

The advantage of DP is the reduced communication overhead, with the cost of affecting the accuracy of the model. For the sake of this work, we opted for SMC in order to more realistically compare the effectiveness of our semi-supervised approach with other approaches, considering privacy as an orthogonal problem. % a common solution since we focused our contribution on the labeled data scarcity problem. 
However, we also plan to investigate how to integrate differential privacy in our framework and its impact on the recognition rate.

%SMC-based solutions enable the third-party cloud service provider to observe only the aggregated result of the clients' updates avoiding that it can get any information from each of those updates. On the other hand, differential privacy techniques consist of adding noise to the outsourced model parameters in order to obscure the involved users characterising information. That noise can be added from the clients before forwarding the locally computed model updates to the server (local differential privacy \cite{geyer2017differentially}) and/or from the server before distributing the shared global model to clients (global differential privacy \cite{mcmahan2017learning}). \\

\subsection{Need of larger datasets for evaluation}

We evaluated \acronym\ choosing those well-known public HAR datasets that involved the highest number of users, simulating the periodicity of FL iterations by partitioning the dataset. However, the effectiveness of \acronym\ on large scale scenarios should be evaluated on significantly larger datasets. By larger, we mean in terms of the number of users involved, the amount of available data for each user, and the number of target activities. Indeed, \acronym\ makes sense when thousands of users are involved, continuously performing activities day after day. However, observing the encouraging results on our limited datasets, we are confident that \acronym\ would perform even better on such large scale evaluations.

Another limitation of the considered datasets is that both of them include data from only one position of the mobile device (trousers pocket). Since mobile devices can actually have different positions (e.g., wristbands), a larger evaluation should also consider different positions of the mobile device.

%Nonetheless, we believe that our method would scale well with respect to the number of users involved, and that the accuracy of the global model would converge to high values after a reasonable number of real-world shards.

\subsection{Resource efficiency}

It is important to mention that \acronym\ is not optimized in terms of computational efficiency.
Indeed, training two deep learning models on mobile devices may be computationally demanding and it may be problematic especially for those devices with low computational capabilities. This problem could be mitigated by relying on trusted edge gateways, as proposed in~\cite{wu2020personalized}.

We want to point out that several research groups are proposing effective ways to dramatically reduce computational efforts for deep learning processes on mobile devices~\cite{lane2016deepx, zhang2019deep}.
Moreover, the GPU modules embedded in recent smartphones exhibit performances similar to the ones of entry-level desktop GPUs and this trend is expected to improve in the next few years~\cite{ignatov2019ai}.

Another limitation of our work is that the label propagation model requires storing the collected feature vectors as a graph. This is clearly not sustainable for a long time on a mobile device. This problem could be solved by imposing a limit on the size of the label propagation graph and periodically deleting old or poorly informative nodes. 

%% file: sections/conclusion.tex
In this work we presented \acronym, a novel semi-supervised federated learning framework for activity recognition on mobile devices. \acronym\ takes into account the data scarcity problem, combining active learning and label propagation to semi-automatically annotate sensor data for each user. 
%Annotated data are used to perform local training in federated learning, in order to share local weights to the cloud server.
%\acronym~ also takes advantage of transfer learning to personalize the local model to boost classification. However, the cloud server only receives model weights from a local model that is not personalized, in order to improve the generalization of the global model over unseen users. 
To the best of our knowledge, \acronym\ is the first application of federated learning to personalized activity recognition that is not based on the assumption that labeled data exists for all participating clients.
Our results show that the combination of active learning and label propagation leads to recognition rates that are close to the ones reached by solutions that rely on fully supervised learning to train the local models.

In future work, 
%we will investigate how to include high-level context data (e.g., semantic location, weather condition, etc) to continuously adapt the federated model based on the current user's context. For instance, if the user is in the gym, she will use and improve a model that is specific for physical exercises. On the other hand, if she is at home, the system would consider a federated model more suitable for smart-home environments. We also 
we plan to investigate how federated clustering can further help improving the non-IID problem for HAR~\cite{briggs2020federated}. Indeed, HAR is more effective when the collaborative model only involves users that are similar between them~\cite{sztyler2017online}. We will study solutions based on federated clustering to automatically group users considering model similarity, creating a dedicated federated model for each cluster. 
Also, we plan to extend \acronym\ to automatically learn features from the unlabeled data stream, following the research direction proposed in~\cite{saeed2020federated}.

%by using a different federated model depending on the profile of each user (e.g., age, fitness, weight).  This approach would require to adapt \acronym~ to automatically group users who share similar characteristics and, at the same time, protecting their privacy.

%% file: main.bbl

\begin{thebibliography}{74}
\ifx \bisbn   \undefined \def \bisbn  #1{ISBN #1}\fi
\ifx \binits  \undefined \def \binits#1{#1}\fi
\ifx \bauthor  \undefined \def \bauthor#1{#1}\fi
\ifx \batitle  \undefined \def \batitle#1{#1}\fi
\ifx \bjtitle  \undefined \def \bjtitle#1{#1}\fi
\ifx \bvolume  \undefined \def \bvolume#1{\textbf{#1}}\fi
\ifx \byear  \undefined \def \byear#1{#1}\fi
\ifx \bissue  \undefined \def \bissue#1{#1}\fi
\ifx \bfpage  \undefined \def \bfpage#1{#1}\fi
\ifx \blpage  \undefined \def \blpage #1{#1}\fi
\ifx \burl  \undefined \def \burl#1{\textsf{#1}}\fi
\ifx \doiurl  \undefined \def \doiurl#1{\url{https://doi.org/#1}}\fi
\ifx \betal  \undefined \def \betal{\textit{et al.}}\fi
\ifx \binstitute  \undefined \def \binstitute#1{#1}\fi
\ifx \binstitutionaled  \undefined \def \binstitutionaled#1{#1}\fi
\ifx \bctitle  \undefined \def \bctitle#1{#1}\fi
\ifx \beditor  \undefined \def \beditor#1{#1}\fi
\ifx \bpublisher  \undefined \def \bpublisher#1{#1}\fi
\ifx \bbtitle  \undefined \def \bbtitle#1{#1}\fi
\ifx \bedition  \undefined \def \bedition#1{#1}\fi
\ifx \bseriesno  \undefined \def \bseriesno#1{#1}\fi
\ifx \blocation  \undefined \def \blocation#1{#1}\fi
\ifx \bsertitle  \undefined \def \bsertitle#1{#1}\fi
\ifx \bsnm \undefined \def \bsnm#1{#1}\fi
\ifx \bsuffix \undefined \def \bsuffix#1{#1}\fi
\ifx \bparticle \undefined \def \bparticle#1{#1}\fi
\ifx \barticle \undefined \def \barticle#1{#1}\fi
\bibcommenthead
\ifx \bconfdate \undefined \def \bconfdate #1{#1}\fi
\ifx \botherref \undefined \def \botherref #1{#1}\fi
\ifx \url \undefined \def \url#1{\textsf{#1}}\fi
\ifx \bchapter \undefined \def \bchapter#1{#1}\fi
\ifx \bbook \undefined \def \bbook#1{#1}\fi
\ifx \bcomment \undefined \def \bcomment#1{#1}\fi
\ifx \oauthor \undefined \def \oauthor#1{#1}\fi
\ifx \citeauthoryear \undefined \def \citeauthoryear#1{#1}\fi
\ifx \endbibitem  \undefined \def \endbibitem {}\fi
\ifx \bconflocation  \undefined \def \bconflocation#1{#1}\fi
\ifx \arxivurl  \undefined \def \arxivurl#1{\textsf{#1}}\fi
\csname PreBibitemsHook\endcsname

%%% 1
\bibitem{lara2013survey}
\begin{barticle}
\bauthor{\bsnm{Lara}, \binits{O.D.}},
\bauthor{\bsnm{Labrador}, \binits{M.A.}}, \betal:
\batitle{A survey on human activity recognition using wearable sensors.}
\bjtitle{IEEE Communications Surveys and Tutorials}
\bvolume{15}(\bissue{3}),
\bfpage{1192}--\blpage{1209}
(\byear{2013})
\end{barticle}
\endbibitem

%%% 2
\bibitem{CookFK13}
\begin{barticle}
\bauthor{\bsnm{Cook}, \binits{D.J.}},
\bauthor{\bsnm{Feuz}, \binits{K.D.}},
\bauthor{\bsnm{Krishnan}, \binits{N.C.}}:
\batitle{Transfer learning for activity recognition: A survey}.
\bjtitle{Knowledge and Information Systems}
\bvolume{36}(\bissue{3}),
\bfpage{537}--\blpage{556}
(\byear{2013})
\end{barticle}
\endbibitem

%%% 3
\bibitem{abdallah2018activity}
\begin{barticle}
\bauthor{\bsnm{Abdallah}, \binits{Z.S.}},
\bauthor{\bsnm{Gaber}, \binits{M.M.}},
\bauthor{\bsnm{Srinivasan}, \binits{B.}},
\bauthor{\bsnm{Krishnaswamy}, \binits{S.}}:
\batitle{Activity recognition with evolving data streams: A review}.
\bjtitle{ACM Computing Surveys (CSUR)}
\bvolume{51}(\bissue{4}),
\bfpage{71}
(\byear{2018})
\end{barticle}
\endbibitem

%%% 4
\bibitem{bettini2015privacy}
\begin{barticle}
\bauthor{\bsnm{Bettini}, \binits{C.}},
\bauthor{\bsnm{Riboni}, \binits{D.}}:
\batitle{Privacy protection in pervasive systems: State of the art and
  technical challenges}.
\bjtitle{Pervasive and Mobile Computing}
\bvolume{17},
\bfpage{159}--\blpage{174}
(\byear{2015})
\end{barticle}
\endbibitem

%%% 5
\bibitem{weiss2012impact}
\begin{bchapter}
\bauthor{\bsnm{Weiss}, \binits{G.M.}},
\bauthor{\bsnm{Lockhart}, \binits{J.}}:
\bctitle{The impact of personalization on smartphone-based activity
  recognition}.
In: \bbtitle{Workshops at the Twenty-Sixth AAAI Conference on Artificial
  Intelligence}
(\byear{2012}).
\bcomment{Citeseer}
\end{bchapter}
\endbibitem

%%% 6
\bibitem{mcmahan2017communication}
\begin{bchapter}
\bauthor{\bsnm{McMahan}, \binits{B.}},
\bauthor{\bsnm{Moore}, \binits{E.}},
\bauthor{\bsnm{Ramage}, \binits{D.}},
\bauthor{\bsnm{Hampson}, \binits{S.}},
\bauthor{\bparticle{y} \bsnm{Arcas}, \binits{B.A.}}:
\bctitle{Communication-efficient learning of deep networks from decentralized
  data}.
In: \bbtitle{Artificial Intelligence and Statistics},
pp. \bfpage{1273}--\blpage{1282}
(\byear{2017}).
\bcomment{PMLR}
\end{bchapter}
\endbibitem

%%% 7
\bibitem{yang2019federated}
\begin{barticle}
\bauthor{\bsnm{Yang}, \binits{Q.}},
\bauthor{\bsnm{Liu}, \binits{Y.}},
\bauthor{\bsnm{Chen}, \binits{T.}},
\bauthor{\bsnm{Tong}, \binits{Y.}}:
\batitle{Federated machine learning: Concept and applications}.
\bjtitle{ACM Transactions on Intelligent Systems and Technology (TIST)}
\bvolume{10}(\bissue{2}),
\bfpage{1}--\blpage{19}
(\byear{2019})
\end{barticle}
\endbibitem

%%% 8
\bibitem{chen2020fedhealth}
\begin{botherref}
\oauthor{\bsnm{Chen}, \binits{Y.}},
\oauthor{\bsnm{Qin}, \binits{X.}},
\oauthor{\bsnm{Wang}, \binits{J.}},
\oauthor{\bsnm{Yu}, \binits{C.}},
\oauthor{\bsnm{Gao}, \binits{W.}}:
Fedhealth: A federated transfer learning framework for wearable healthcare.
IEEE Intelligent Systems
(2020)
\end{botherref}
\endbibitem

%%% 9
\bibitem{kairouz2019advances}
\begin{botherref}
\oauthor{\bsnm{Kairouz}, \binits{P.}},
\oauthor{\bsnm{McMahan}, \binits{H.B.}},
\oauthor{\bsnm{Avent}, \binits{B.}},
\oauthor{\bsnm{Bellet}, \binits{A.}},
\oauthor{\bsnm{Bennis}, \binits{M.}},
\oauthor{\bsnm{Bhagoji}, \binits{A.N.}},
\oauthor{\bsnm{Bonawitz}, \binits{K.}},
\oauthor{\bsnm{Charles}, \binits{Z.}},
\oauthor{\bsnm{Cormode}, \binits{G.}},
\oauthor{\bsnm{Cummings}, \binits{R.}}, et al.:
Advances and open problems in federated learning.
arXiv preprint arXiv:1912.04977
(2019)
\end{botherref}
\endbibitem

%%% 10
\bibitem{hard2018federated}
\begin{botherref}
\oauthor{\bsnm{Hard}, \binits{A.}},
\oauthor{\bsnm{Rao}, \binits{K.}},
\oauthor{\bsnm{Mathews}, \binits{R.}},
\oauthor{\bsnm{Ramaswamy}, \binits{S.}},
\oauthor{\bsnm{Beaufays}, \binits{F.}},
\oauthor{\bsnm{Augenstein}, \binits{S.}},
\oauthor{\bsnm{Eichner}, \binits{H.}},
\oauthor{\bsnm{Kiddon}, \binits{C.}},
\oauthor{\bsnm{Ramage}, \binits{D.}}:
Federated learning for mobile keyboard prediction.
arXiv preprint arXiv:1811.03604
(2018)
\end{botherref}
\endbibitem

%%% 11
\bibitem{ek2020evaluation}
\begin{bchapter}
\bauthor{\bsnm{Ek}, \binits{S.}},
\bauthor{\bsnm{Portet}, \binits{F.}},
\bauthor{\bsnm{Lalanda}, \binits{P.}},
\bauthor{\bsnm{Vega}, \binits{G.}}:
\bctitle{Evaluation of federated learning aggregation algorithms: application
  to human activity recognition}.
In: \bbtitle{Adjunct Proceedings of the 2020 ACM International Joint Conference
  on Pervasive and Ubiquitous Computing and Proceedings of the 2020 ACM
  International Symposium on Wearable Computers},
pp. \bfpage{638}--\blpage{643}
(\byear{2020})
\end{bchapter}
\endbibitem

%%% 12
\bibitem{kwapisz2011activity}
\begin{barticle}
\bauthor{\bsnm{Kwapisz}, \binits{J.R.}},
\bauthor{\bsnm{Weiss}, \binits{G.M.}},
\bauthor{\bsnm{Moore}, \binits{S.A.}}:
\batitle{Activity recognition using cell phone accelerometers}.
\bjtitle{ACM SigKDD Explorations Newsletter}
\bvolume{12}(\bissue{2}),
\bfpage{74}--\blpage{82}
(\byear{2011})
\end{barticle}
\endbibitem

%%% 13
\bibitem{gyorbiro2009activity}
\begin{barticle}
\bauthor{\bsnm{Gy{\"o}rb{\'\i}r{\'o}}, \binits{N.}},
\bauthor{\bsnm{F{\'a}bi{\'a}n}, \binits{{\'A}.}},
\bauthor{\bsnm{Hom{\'a}nyi}, \binits{G.}}:
\batitle{An activity recognition system for mobile phones}.
\bjtitle{Mobile Networks and Applications}
\bvolume{14}(\bissue{1}),
\bfpage{82}--\blpage{91}
(\byear{2009})
\end{barticle}
\endbibitem

%%% 14
\bibitem{sun2010activity}
\begin{bchapter}
\bauthor{\bsnm{Sun}, \binits{L.}},
\bauthor{\bsnm{Zhang}, \binits{D.}},
\bauthor{\bsnm{Li}, \binits{B.}},
\bauthor{\bsnm{Guo}, \binits{B.}},
\bauthor{\bsnm{Li}, \binits{S.}}:
\bctitle{Activity recognition on an accelerometer embedded mobile phone with
  varying positions and orientations}.
In: \bbtitle{International Conference on Ubiquitous Intelligence and
  Computing},
pp. \bfpage{548}--\blpage{562}
(\byear{2010}).
\bcomment{Springer}
\end{bchapter}
\endbibitem

%%% 15
\bibitem{bao2004activity}
\begin{bchapter}
\bauthor{\bsnm{Bao}, \binits{L.}},
\bauthor{\bsnm{Intille}, \binits{S.S.}}:
\bctitle{Activity recognition from user-annotated acceleration data}.
In: \bbtitle{Pervasive Computing: Second International Conference, PERVASIVE
  2004, Linz/Vienna, Austria, April 21-23, 2004. Proceedings},
pp. \bfpage{1}--\blpage{17}.
\bpublisher{Springer},
\blocation{Berlin, Heidelberg}
(\byear{2004})
\end{bchapter}
\endbibitem

%%% 16
\bibitem{BullingBS14}
\begin{barticle}
\bauthor{\bsnm{Bulling}, \binits{A.}},
\bauthor{\bsnm{Blanke}, \binits{U.}},
\bauthor{\bsnm{Schiele}, \binits{B.}}:
\batitle{A tutorial on human activity recognition using body-worn inertial
  sensors}.
\bjtitle{{ACM} Computing Surveys}
\bvolume{46}(\bissue{3}),
\bfpage{33}--\blpage{13333}
(\byear{2014})
\end{barticle}
\endbibitem

%%% 17
\bibitem{kwon2014unsupervised}
\begin{barticle}
\bauthor{\bsnm{Kwon}, \binits{Y.}},
\bauthor{\bsnm{Kang}, \binits{K.}},
\bauthor{\bsnm{Bae}, \binits{C.}}:
\batitle{Unsupervised learning for human activity recognition using smartphone
  sensors}.
\bjtitle{Expert Systems with Applications}
\bvolume{41}(\bissue{14}),
\bfpage{6067}--\blpage{6074}
(\byear{2014})
\end{barticle}
\endbibitem

%%% 18
\bibitem{chen2009ontology}
\begin{botherref}
\oauthor{\bsnm{Chen}, \binits{L.}},
\oauthor{\bsnm{Nugent}, \binits{C.}}:
Ontology-based activity recognition in intelligent pervasive environments.
International Journal of Web Information Systems
(2009)
\end{botherref}
\endbibitem

%%% 19
\bibitem{civitarese2019polaris}
\begin{barticle}
\bauthor{\bsnm{Civitarese}, \binits{G.}},
\bauthor{\bsnm{Sztyler}, \binits{T.}},
\bauthor{\bsnm{Riboni}, \binits{D.}},
\bauthor{\bsnm{Bettini}, \binits{C.}},
\bauthor{\bsnm{Stuckenschmidt}, \binits{H.}}:
\batitle{Polaris: Probabilistic and ontological activity recognition in
  smart-homes}.
\bjtitle{IEEE Transactions on Knowledge and Data Engineering}
\bvolume{33}(\bissue{1}),
\bfpage{209}--\blpage{223}
(\byear{2019})
\end{barticle}
\endbibitem

%%% 20
\bibitem{chawla2002smote}
\begin{barticle}
\bauthor{\bsnm{Chawla}, \binits{N.V.}},
\bauthor{\bsnm{Bowyer}, \binits{K.W.}},
\bauthor{\bsnm{Hall}, \binits{L.O.}},
\bauthor{\bsnm{Kegelmeyer}, \binits{W.P.}}:
\batitle{Smote: synthetic minority over-sampling technique}.
\bjtitle{Journal of artificial intelligence research}
\bvolume{16},
\bfpage{321}--\blpage{357}
(\byear{2002})
\end{barticle}
\endbibitem

%%% 21
\bibitem{rashid2019times}
\begin{barticle}
\bauthor{\bsnm{Rashid}, \binits{K.M.}},
\bauthor{\bsnm{Louis}, \binits{J.}}:
\batitle{Times-series data augmentation and deep learning for construction
  equipment activity recognition}.
\bjtitle{Advanced Engineering Informatics}
\bvolume{42},
\bfpage{100944}
(\byear{2019})
\end{barticle}
\endbibitem

%%% 22
\bibitem{wang2018sensorygans}
\begin{bchapter}
\bauthor{\bsnm{Wang}, \binits{J.}},
\bauthor{\bsnm{Chen}, \binits{Y.}},
\bauthor{\bsnm{Gu}, \binits{Y.}},
\bauthor{\bsnm{Xiao}, \binits{Y.}},
\bauthor{\bsnm{Pan}, \binits{H.}}:
\bctitle{Sensorygans: An effective generative adversarial framework for
  sensor-based human activity recognition}.
In: \bbtitle{2018 International Joint Conference on Neural Networks (IJCNN)},
pp. \bfpage{1}--\blpage{8}
(\byear{2018}).
\bcomment{IEEE}
\end{bchapter}
\endbibitem

%%% 23
\bibitem{chan2020unified}
\begin{botherref}
\oauthor{\bsnm{Chan}, \binits{M.H.}},
\oauthor{\bsnm{Noor}, \binits{M.H.M.}}:
A unified generative model using generative adversarial network for activity
  recognition.
Journal of Ambient Intelligence and Humanized Computing,
1--10
(2020)
\end{botherref}
\endbibitem

%%% 24
\bibitem{cook2013transfer}
\begin{barticle}
\bauthor{\bsnm{Cook}, \binits{D.}},
\bauthor{\bsnm{Feuz}, \binits{K.D.}},
\bauthor{\bsnm{Krishnan}, \binits{N.C.}}:
\batitle{Transfer learning for activity recognition: A survey}.
\bjtitle{Knowledge and information systems}
\bvolume{36}(\bissue{3}),
\bfpage{537}--\blpage{556}
(\byear{2013})
\end{barticle}
\endbibitem

%%% 25
\bibitem{wang2018deep}
\begin{bchapter}
\bauthor{\bsnm{Wang}, \binits{J.}},
\bauthor{\bsnm{Zheng}, \binits{V.W.}},
\bauthor{\bsnm{Chen}, \binits{Y.}},
\bauthor{\bsnm{Huang}, \binits{M.}}:
\bctitle{Deep transfer learning for cross-domain activity recognition}.
In: \bbtitle{Proceedings of the 3rd International Conference on Crowd Science
  and Engineering},
pp. \bfpage{1}--\blpage{8}
(\byear{2018})
\end{bchapter}
\endbibitem

%%% 26
\bibitem{sanabria2021unsupervised}
\begin{barticle}
\bauthor{\bsnm{Sanabria}, \binits{A.R.}},
\bauthor{\bsnm{Zambonelli}, \binits{F.}},
\bauthor{\bsnm{Ye}, \binits{J.}}:
\batitle{Unsupervised domain adaptation in activity recognition: A gan-based
  approach}.
\bjtitle{IEEE Access}
\bvolume{9},
\bfpage{19421}--\blpage{19438}
(\byear{2021})
\end{barticle}
\endbibitem

%%% 27
\bibitem{soleimani2021cross}
\begin{barticle}
\bauthor{\bsnm{Soleimani}, \binits{E.}},
\bauthor{\bsnm{Nazerfard}, \binits{E.}}:
\batitle{Cross-subject transfer learning in human activity recognition systems
  using generative adversarial networks}.
\bjtitle{Neurocomputing}
\bvolume{426},
\bfpage{26}--\blpage{34}
(\byear{2021})
\end{barticle}
\endbibitem

%%% 28
\bibitem{stikic2008exploring}
\begin{bchapter}
\bauthor{\bsnm{Stikic}, \binits{M.}},
\bauthor{\bsnm{Van~Laerhoven}, \binits{K.}},
\bauthor{\bsnm{Schiele}, \binits{B.}}:
\bctitle{Exploring semi-supervised and active learning for activity
  recognition}.
In: \bbtitle{2008 12th IEEE International Symposium on Wearable Computers},
pp. \bfpage{81}--\blpage{88}
(\byear{2008}).
\bcomment{IEEE}
\end{bchapter}
\endbibitem

%%% 29
\bibitem{guan2007activity}
\begin{bchapter}
\bauthor{\bsnm{Guan}, \binits{D.}},
\bauthor{\bsnm{Yuan}, \binits{W.}},
\bauthor{\bsnm{Lee}, \binits{Y.-K.}},
\bauthor{\bsnm{Gavrilov}, \binits{A.}},
\bauthor{\bsnm{Lee}, \binits{S.}}:
\bctitle{Activity recognition based on semi-supervised learning}.
In: \bbtitle{Embedded and Real-Time Computing Systems and Applications, 2007.
  RTCSA 2007. 13th IEEE International Conference On},
pp. \bfpage{469}--\blpage{475}
(\byear{2007}).
\bcomment{IEEE}
\end{bchapter}
\endbibitem

%%% 30
\bibitem{longstaff2010improving}
\begin{bchapter}
\bauthor{\bsnm{Longstaff}, \binits{B.}},
\bauthor{\bsnm{Reddy}, \binits{S.}},
\bauthor{\bsnm{Estrin}, \binits{D.}}:
\bctitle{Improving activity classification for health applications on mobile
  devices using active and semi-supervised learning}.
In: \bbtitle{Pervasive Computing Technologies for Healthcare (PervasiveHealth),
  2010 4th International Conference on Pervasive Computing Technologies for
  Healthcare},
pp. \bfpage{1}--\blpage{7}
(\byear{2010}).
\bcomment{IEEE}
\end{bchapter}
\endbibitem

%%% 31
\bibitem{stikic2009multi}
\begin{bchapter}
\bauthor{\bsnm{Stikic}, \binits{M.}},
\bauthor{\bsnm{Larlus}, \binits{D.}},
\bauthor{\bsnm{Schiele}, \binits{B.}}:
\bctitle{Multi-graph based semi-supervised learning for activity recognition}.
In: \bbtitle{2009 International Symposium on Wearable Computers},
pp. \bfpage{85}--\blpage{92}
(\byear{2009}).
\bcomment{IEEE}
\end{bchapter}
\endbibitem

%%% 32
\bibitem{lee2014activity}
\begin{barticle}
\bauthor{\bsnm{Lee}, \binits{Y.-S.}},
\bauthor{\bsnm{Cho}, \binits{S.-B.}}:
\batitle{Activity recognition with android phone using mixture-of-experts
  co-trained with labeled and unlabeled data}.
\bjtitle{Neurocomputing}
\bvolume{126},
\bfpage{106}--\blpage{115}
(\byear{2014})
\end{barticle}
\endbibitem

%%% 33
\bibitem{miu2015bootstrapping}
\begin{bchapter}
\bauthor{\bsnm{Miu}, \binits{T.}},
\bauthor{\bsnm{Missier}, \binits{P.}},
\bauthor{\bsnm{Pl{\"o}tz}, \binits{T.}}:
\bctitle{Bootstrapping personalised human activity recognition models using
  online active learning}.
In: \bbtitle{2015 IEEE International Conference on Computer and Information
  Technology; Ubiquitous Computing and Communications; Dependable, Autonomic
  and Secure Computing; Pervasive Intelligence and Computing},
pp. \bfpage{1138}--\blpage{1147}
(\byear{2015}).
\bcomment{IEEE}
\end{bchapter}
\endbibitem

%%% 34
\bibitem{abdallah2015adaptive}
\begin{barticle}
\bauthor{\bsnm{Abdallah}, \binits{Z.S.}},
\bauthor{\bsnm{Gaber}, \binits{M.M.}},
\bauthor{\bsnm{Srinivasan}, \binits{B.}},
\bauthor{\bsnm{Krishnaswamy}, \binits{S.}}:
\batitle{Adaptive mobile activity recognition system with evolving data
  streams}.
\bjtitle{Neurocomputing}
\bvolume{150},
\bfpage{304}--\blpage{317}
(\byear{2015})
\end{barticle}
\endbibitem

%%% 35
\bibitem{hossain2017active}
\begin{barticle}
\bauthor{\bsnm{Hossain}, \binits{H.S.}},
\bauthor{\bsnm{Khan}, \binits{M.A.A.H.}},
\bauthor{\bsnm{Roy}, \binits{N.}}:
\batitle{Active learning enabled activity recognition}.
\bjtitle{Pervasive and Mobile Computing}
\bvolume{38},
\bfpage{312}--\blpage{330}
(\byear{2017})
\end{barticle}
\endbibitem

%%% 36
\bibitem{nguyen2018dealing}
\begin{bchapter}
\bauthor{\bsnm{Nguyen}, \binits{K.T.}},
\bauthor{\bsnm{Portet}, \binits{F.}},
\bauthor{\bsnm{Garbay}, \binits{C.}}:
\bctitle{Dealing with imbalanced data sets for human activity recognition using
  mobile phone sensors}.
In: \bbtitle{3rd International Workshop on Smart Sensing Systems}
(\byear{2018})
\end{bchapter}
\endbibitem

%%% 37
\bibitem{bettini2020caviar}
\begin{barticle}
\bauthor{\bsnm{Bettini}, \binits{C.}},
\bauthor{\bsnm{Civitarese}, \binits{G.}},
\bauthor{\bsnm{Presotto}, \binits{R.}}:
\batitle{Caviar: Context-driven active and incremental activity recognition}.
\bjtitle{Knowledge-Based Systems}
\bvolume{196},
\bfpage{105816}
(\byear{2020})
\end{barticle}
\endbibitem

%%% 38
\bibitem{voigt2017eu}
\begin{botherref}
\oauthor{\bsnm{Voigt}, \binits{P.}},
\oauthor{\bparticle{Von~dem} \bsnm{Bussche}, \binits{A.}}:
The eu general data protection regulation (gdpr).
A Practical Guide, 1st Ed., Cham: Springer International Publishing
(2017)
\end{botherref}
\endbibitem

%%% 39
\bibitem{samarati2014data}
\begin{bchapter}
\bauthor{\bsnm{Samarati}, \binits{P.}}:
\bctitle{Data security and privacy in the cloud}.
In: \bbtitle{International Conference on Information Security Practice and
  Experience},
pp. \bfpage{28}--\blpage{41}
(\byear{2014}).
\bcomment{Springer}
\end{bchapter}
\endbibitem

%%% 40
\bibitem{konevcny2016federated}
\begin{botherref}
\oauthor{\bsnm{Kone{\v{c}}n{\`y}}, \binits{J.}},
\oauthor{\bsnm{McMahan}, \binits{H.B.}},
\oauthor{\bsnm{Yu}, \binits{F.X.}},
\oauthor{\bsnm{Richt{\'a}rik}, \binits{P.}},
\oauthor{\bsnm{Suresh}, \binits{A.T.}},
\oauthor{\bsnm{Bacon}, \binits{D.}}:
Federated learning: Strategies for improving communication efficiency.
arXiv preprint arXiv:1610.05492
(2016)
\end{botherref}
\endbibitem

%%% 41
\bibitem{bonawitz2017practical}
\begin{bchapter}
\bauthor{\bsnm{Bonawitz}, \binits{K.}},
\bauthor{\bsnm{Ivanov}, \binits{V.}},
\bauthor{\bsnm{Kreuter}, \binits{B.}},
\bauthor{\bsnm{Marcedone}, \binits{A.}},
\bauthor{\bsnm{McMahan}, \binits{H.B.}},
\bauthor{\bsnm{Patel}, \binits{S.}},
\bauthor{\bsnm{Ramage}, \binits{D.}},
\bauthor{\bsnm{Segal}, \binits{A.}},
\bauthor{\bsnm{Seth}, \binits{K.}}:
\bctitle{Practical secure aggregation for privacy-preserving machine learning}.
In: \bbtitle{Proceedings of the 2017 ACM SIGSAC Conference on Computer and
  Communications Security},
pp. \bfpage{1175}--\blpage{1191}
(\byear{2017})
\end{bchapter}
\endbibitem

%%% 42
\bibitem{damaskinos2020fleet}
\begin{bchapter}
\bauthor{\bsnm{Damaskinos}, \binits{G.}},
\bauthor{\bsnm{Guerraoui}, \binits{R.}},
\bauthor{\bsnm{Kermarrec}, \binits{A.-M.}},
\bauthor{\bsnm{Nitu}, \binits{V.}},
\bauthor{\bsnm{Patra}, \binits{R.}},
\bauthor{\bsnm{Taiani}, \binits{F.}}:
\bctitle{Fleet: Online federated learning via staleness awareness and
  performance prediction}.
In: \bbtitle{Proceedings of the 21st International Middleware Conference},
pp. \bfpage{163}--\blpage{177}
(\byear{2020})
\end{bchapter}
\endbibitem

%%% 43
\bibitem{fallah2020personalized}
\begin{botherref}
\oauthor{\bsnm{Fallah}, \binits{A.}},
\oauthor{\bsnm{Mokhtari}, \binits{A.}},
\oauthor{\bsnm{Ozdaglar}, \binits{A.}}:
Personalized federated learning: A meta-learning approach.
arXiv preprint arXiv:2002.07948
(2020)
\end{botherref}
\endbibitem

%%% 44
\bibitem{ek2021federated}
\begin{bchapter}
\bauthor{\bsnm{Ek}, \binits{S.}},
\bauthor{\bsnm{Portet}, \binits{F.}},
\bauthor{\bsnm{Lalanda}, \binits{P.}},
\bauthor{\bsnm{Vega}, \binits{G.}}:
\bctitle{A federated learning aggregation algorithm for pervasive computing:
  Evaluation and comparison}.
In: \bbtitle{19th IEEE International Conference on Pervasive Computing and
  Communications PerCom 2021}
(\byear{2021})
\end{bchapter}
\endbibitem

%%% 45
\bibitem{sozinov2018human}
\begin{bchapter}
\bauthor{\bsnm{Sozinov}, \binits{K.}},
\bauthor{\bsnm{Vlassov}, \binits{V.}},
\bauthor{\bsnm{Girdzijauskas}, \binits{S.}}:
\bctitle{Human activity recognition using federated learning}.
In: \bbtitle{2018 IEEE Intl Conf on Parallel \& Distributed Processing with
  Applications, Ubiquitous Computing \& Communications, Big Data \& Cloud
  Computing, Social Computing \& Networking, Sustainable Computing \&
  Communications (ISPA/IUCC/BDCloud/SocialCom/SustainCom)},
pp. \bfpage{1103}--\blpage{1111}
(\byear{2018}).
\bcomment{IEEE}
\end{bchapter}
\endbibitem

%%% 46
\bibitem{wu2020personalized}
\begin{botherref}
\oauthor{\bsnm{Wu}, \binits{Q.}},
\oauthor{\bsnm{He}, \binits{K.}},
\oauthor{\bsnm{Chen}, \binits{X.}}:
Personalized federated learning for intelligent iot applications: A cloud-edge
  based framework.
IEEE Computer Graphics and Applications
(2020)
\end{botherref}
\endbibitem

%%% 47
\bibitem{zhao2020semi}
\begin{botherref}
\oauthor{\bsnm{Zhao}, \binits{Y.}},
\oauthor{\bsnm{Liu}, \binits{H.}},
\oauthor{\bsnm{Li}, \binits{H.}},
\oauthor{\bsnm{Barnaghi}, \binits{P.}},
\oauthor{\bsnm{Haddadi}, \binits{H.}}:
Semi-supervised federated learning for activity recognition.
arXiv preprint arXiv:2011.00851
(2020)
\end{botherref}
\endbibitem

%%% 48
\bibitem{wu2020fedhome}
\begin{botherref}
\oauthor{\bsnm{Wu}, \binits{Q.}},
\oauthor{\bsnm{Chen}, \binits{X.}},
\oauthor{\bsnm{Zhou}, \binits{Z.}},
\oauthor{\bsnm{Zhang}, \binits{J.}}:
Fedhome: Cloud-edge based personalized federated learning for in-home health
  monitoring.
IEEE Transactions on Mobile Computing
(2020)
\end{botherref}
\endbibitem

%%% 49
\bibitem{lee2021opportunistic}
\begin{bchapter}
\bauthor{\bsnm{Lee}, \binits{S.}},
\bauthor{\bsnm{Zheng}, \binits{X.}},
\bauthor{\bsnm{Hua}, \binits{J.}},
\bauthor{\bsnm{Vikalo}, \binits{H.}},
\bauthor{\bsnm{Julien}, \binits{C.}}:
\bctitle{Opportunistic federated learning: An exploration of egocentric
  collaboration for pervasive computing applications}.
In: \bbtitle{2021 IEEE International Conference on Pervasive Computing and
  Communications (PerCom)},
pp. \bfpage{1}--\blpage{8}
(\byear{2021}).
\bcomment{IEEE}
\end{bchapter}
\endbibitem

%%% 50
\bibitem{kelli2021ids}
\begin{barticle}
\bauthor{\bsnm{Kelli}, \binits{V.}},
\bauthor{\bsnm{Argyriou}, \binits{V.}},
\bauthor{\bsnm{Lagkas}, \binits{T.}},
\bauthor{\bsnm{Fragulis}, \binits{G.}},
\bauthor{\bsnm{Grigoriou}, \binits{E.}},
\bauthor{\bsnm{Sarigiannidis}, \binits{P.}}:
\batitle{Ids for industrial applications: a federated learning approach with
  active personalization}.
\bjtitle{Sensors}
\bvolume{21}(\bissue{20}),
\bfpage{6743}
(\byear{2021})
\end{barticle}
\endbibitem

%%% 51
\bibitem{saeed2020federated}
\begin{barticle}
\bauthor{\bsnm{Saeed}, \binits{A.}},
\bauthor{\bsnm{Salim}, \binits{F.D.}},
\bauthor{\bsnm{Ozcelebi}, \binits{T.}},
\bauthor{\bsnm{Lukkien}, \binits{J.}}:
\batitle{Federated self-supervised learning of multisensor representations for
  embedded intelligence}.
\bjtitle{IEEE Internet of Things Journal}
\bvolume{8}(\bissue{2}),
\bfpage{1030}--\blpage{1040}
(\byear{2020})
\end{barticle}
\endbibitem

%%% 52
\bibitem{yu2021fedhar}
\begin{botherref}
\oauthor{\bsnm{Yu}, \binits{H.}},
\oauthor{\bsnm{Chen}, \binits{Z.}},
\oauthor{\bsnm{Zhang}, \binits{X.}},
\oauthor{\bsnm{Chen}, \binits{X.}},
\oauthor{\bsnm{Zhuang}, \binits{F.}},
\oauthor{\bsnm{Xiong}, \binits{H.}},
\oauthor{\bsnm{Cheng}, \binits{X.}}:
Fedhar: Semi-supervised online learning for personalized federated human
  activity recognition.
IEEE Transactions on Mobile Computing
(2021)
\end{botherref}
\endbibitem

%%% 53
\bibitem{arivazhagan2019federated}
\begin{botherref}
\oauthor{\bsnm{Arivazhagan}, \binits{M.G.}},
\oauthor{\bsnm{Aggarwal}, \binits{V.}},
\oauthor{\bsnm{Singh}, \binits{A.K.}},
\oauthor{\bsnm{Choudhary}, \binits{S.}}:
Federated learning with personalization layers.
arXiv preprint arXiv:1912.00818
(2019)
\end{botherref}
\endbibitem

%%% 54
\bibitem{nasr2019comprehensive}
\begin{bchapter}
\bauthor{\bsnm{Nasr}, \binits{M.}},
\bauthor{\bsnm{Shokri}, \binits{R.}},
\bauthor{\bsnm{Houmansadr}, \binits{A.}}:
\bctitle{Comprehensive privacy analysis of deep learning: Passive and active
  white-box inference attacks against centralized and federated learning}.
In: \bbtitle{2019 IEEE Symposium on Security and Privacy (SP)},
pp. \bfpage{739}--\blpage{753}
(\byear{2019}).
\bcomment{IEEE}
\end{bchapter}
\endbibitem

%%% 55
\bibitem{cruciani2019comparing}
\begin{bchapter}
\bauthor{\bsnm{Cruciani}, \binits{F.}},
\bauthor{\bsnm{Vafeiadis}, \binits{A.}},
\bauthor{\bsnm{Nugent}, \binits{C.}},
\bauthor{\bsnm{Cleland}, \binits{I.}},
\bauthor{\bsnm{McCullagh}, \binits{P.}},
\bauthor{\bsnm{Votis}, \binits{K.}},
\bauthor{\bsnm{Giakoumis}, \binits{D.}},
\bauthor{\bsnm{Tzovaras}, \binits{D.}},
\bauthor{\bsnm{Chen}, \binits{L.}},
\bauthor{\bsnm{Hamzaoui}, \binits{R.}}:
\bctitle{Comparing cnn and human crafted features for human activity
  recognition}.
In: \bbtitle{2019 IEEE SmartWorld, Ubiquitous Intelligence \& Computing,
  Advanced \& Trusted Computing, Scalable Computing \& Communications, Cloud \&
  Big Data Computing, Internet of People and Smart City Innovation
  (SmartWorld/SCALCOM/UIC/ATC/CBDCom/IOP/SCI)},
pp. \bfpage{960}--\blpage{967}
(\byear{2019}).
\bcomment{IEEE}
\end{bchapter}
\endbibitem

%%% 56
\bibitem{yosinski2014transferable}
\begin{bchapter}
\bauthor{\bsnm{Yosinski}, \binits{J.}},
\bauthor{\bsnm{Clune}, \binits{J.}},
\bauthor{\bsnm{Bengio}, \binits{Y.}},
\bauthor{\bsnm{Lipson}, \binits{H.}}:
\bctitle{How transferable are features in deep neural networks?}
In: \bbtitle{Advances in Neural Information Processing Systems},
pp. \bfpage{3320}--\blpage{3328}
(\byear{2014})
\end{bchapter}
\endbibitem

%%% 57
\bibitem{vzliobaite2013active}
\begin{barticle}
\bauthor{\bsnm{{\v{Z}}liobait{\.e}}, \binits{I.}},
\bauthor{\bsnm{Bifet}, \binits{A.}},
\bauthor{\bsnm{Pfahringer}, \binits{B.}},
\bauthor{\bsnm{Holmes}, \binits{G.}}:
\batitle{Active learning with drifting streaming data}.
\bjtitle{IEEE transactions on neural networks and learning systems}
\bvolume{25}(\bissue{1}),
\bfpage{27}--\blpage{39}
(\byear{2013})
\end{barticle}
\endbibitem

%%% 58
\bibitem{zhou2004learning}
\begin{barticle}
\bauthor{\bsnm{Zhou}, \binits{D.}},
\bauthor{\bsnm{Bousquet}, \binits{O.}},
\bauthor{\bsnm{Lal}, \binits{T.N.}},
\bauthor{\bsnm{Weston}, \binits{J.}},
\bauthor{\bsnm{Sch{\"o}lkopf}, \binits{B.}}:
\batitle{Learning with local and global consistency}.
\bjtitle{Advances in neural information processing systems}
\bvolume{16}(\bissue{16}),
\bfpage{321}--\blpage{328}
(\byear{2004})
\end{barticle}
\endbibitem

%%% 59
\bibitem{widmann2017graph}
\begin{bchapter}
\bauthor{\bsnm{Widmann}, \binits{N.}},
\bauthor{\bsnm{Verberne}, \binits{S.}}:
\bctitle{Graph-based semi-supervised learning for text classification}.
In: \bbtitle{Proceedings of the ACM SIGIR International Conference on Theory of
  Information Retrieval},
pp. \bfpage{59}--\blpage{66}
(\byear{2017})
\end{bchapter}
\endbibitem

%%% 60
\bibitem{vavoulas2016mobiact}
\begin{bchapter}
\bauthor{\bsnm{Vavoulas}, \binits{G.}},
\bauthor{\bsnm{Chatzaki}, \binits{C.}},
\bauthor{\bsnm{Malliotakis}, \binits{T.}},
\bauthor{\bsnm{Pediaditis}, \binits{M.}},
\bauthor{\bsnm{Tsiknakis}, \binits{M.}}:
\bctitle{The mobiact dataset: Recognition of activities of daily living using
  smartphones.}
In: \bbtitle{ICT4AgeingWell},
pp. \bfpage{143}--\blpage{151}
(\byear{2016})
\end{bchapter}
\endbibitem

%%% 61
\bibitem{kingma2017adam}
\begin{botherref}
\oauthor{\bsnm{Kingma}, \binits{D.P.}},
\oauthor{\bsnm{Ba}, \binits{J.}}:
Adam: A Method for Stochastic Optimization
(2017)
\end{botherref}
\endbibitem

%%% 62
\bibitem{wan2020deep}
\begin{barticle}
\bauthor{\bsnm{Wan}, \binits{S.}},
\bauthor{\bsnm{Qi}, \binits{L.}},
\bauthor{\bsnm{Xu}, \binits{X.}},
\bauthor{\bsnm{Tong}, \binits{C.}},
\bauthor{\bsnm{Gu}, \binits{Z.}}:
\batitle{Deep learning models for real-time human activity recognition with
  smartphones}.
\bjtitle{Mobile Networks and Applications}
\bvolume{25}(\bissue{2}),
\bfpage{743}--\blpage{755}
(\byear{2020})
\end{barticle}
\endbibitem

%%% 63
\bibitem{nasr2018comprehensive}
\begin{botherref}
\oauthor{\bsnm{Nasr}, \binits{M.}},
\oauthor{\bsnm{Shokri}, \binits{R.}},
\oauthor{\bsnm{Houmansadr}, \binits{A.}}:
Comprehensive privacy analysis of deep learning: Stand-alone and federated
  learning under passive and active white-box inference attacks.
arXiv preprint arXiv:1812.00910
(2018)
\end{botherref}
\endbibitem

%%% 64
\bibitem{shokri2017membership}
\begin{bchapter}
\bauthor{\bsnm{Shokri}, \binits{R.}},
\bauthor{\bsnm{Stronati}, \binits{M.}},
\bauthor{\bsnm{Song}, \binits{C.}},
\bauthor{\bsnm{Shmatikov}, \binits{V.}}:
\bctitle{Membership inference attacks against machine learning models}.
In: \bbtitle{2017 IEEE Symposium on Security and Privacy (SP)},
pp. \bfpage{3}--\blpage{18}
(\byear{2017}).
\bcomment{IEEE}
\end{bchapter}
\endbibitem

%%% 65
\bibitem{bhowmick2018protection}
\begin{botherref}
\oauthor{\bsnm{Bhowmick}, \binits{A.}},
\oauthor{\bsnm{Duchi}, \binits{J.}},
\oauthor{\bsnm{Freudiger}, \binits{J.}},
\oauthor{\bsnm{Kapoor}, \binits{G.}},
\oauthor{\bsnm{Rogers}, \binits{R.}}:
Protection against reconstruction and its applications in private federated
  learning.
arXiv preprint arXiv:1812.00984
(2018)
\end{botherref}
\endbibitem

%%% 66
\bibitem{cramer2000general}
\begin{bchapter}
\bauthor{\bsnm{Cramer}, \binits{R.}},
\bauthor{\bsnm{Damg{\aa}rd}, \binits{I.}},
\bauthor{\bsnm{Maurer}, \binits{U.}}:
\bctitle{General secure multi-party computation from any linear secret-sharing
  scheme}.
In: \bbtitle{International Conference on the Theory and Applications of
  Cryptographic Techniques},
pp. \bfpage{316}--\blpage{334}
(\byear{2000}).
\bcomment{Springer}
\end{bchapter}
\endbibitem

%%% 67
\bibitem{agarwal2018cpsgd}
\begin{bchapter}
\bauthor{\bsnm{Agarwal}, \binits{N.}},
\bauthor{\bsnm{Suresh}, \binits{A.T.}},
\bauthor{\bsnm{Yu}, \binits{F.X.X.}},
\bauthor{\bsnm{Kumar}, \binits{S.}},
\bauthor{\bsnm{McMahan}, \binits{B.}}:
\bctitle{cpsgd: Communication-efficient and differentially-private distributed
  sgd}.
In: \bbtitle{Advances in Neural Information Processing Systems},
pp. \bfpage{7564}--\blpage{7575}
(\byear{2018})
\end{bchapter}
\endbibitem

%%% 68
\bibitem{dwork2008differential}
\begin{bchapter}
\bauthor{\bsnm{Dwork}, \binits{C.}}:
\bctitle{Differential privacy: A survey of results}.
In: \bbtitle{International Conference on Theory and Applications of Models of
  Computation},
pp. \bfpage{1}--\blpage{19}
(\byear{2008}).
\bcomment{Springer}
\end{bchapter}
\endbibitem

%%% 69
\bibitem{truex2019hybrid}
\begin{bchapter}
\bauthor{\bsnm{Truex}, \binits{S.}},
\bauthor{\bsnm{Baracaldo}, \binits{N.}},
\bauthor{\bsnm{Anwar}, \binits{A.}},
\bauthor{\bsnm{Steinke}, \binits{T.}},
\bauthor{\bsnm{Ludwig}, \binits{H.}},
\bauthor{\bsnm{Zhang}, \binits{R.}},
\bauthor{\bsnm{Zhou}, \binits{Y.}}:
\bctitle{A hybrid approach to privacy-preserving federated learning}.
In: \bbtitle{Proceedings of the 12th ACM Workshop on Artificial Intelligence
  and Security},
pp. \bfpage{1}--\blpage{11}
(\byear{2019})
\end{bchapter}
\endbibitem

%%% 70
\bibitem{lane2016deepx}
\begin{bchapter}
\bauthor{\bsnm{Lane}, \binits{N.D.}},
\bauthor{\bsnm{Bhattacharya}, \binits{S.}},
\bauthor{\bsnm{Georgiev}, \binits{P.}},
\bauthor{\bsnm{Forlivesi}, \binits{C.}},
\bauthor{\bsnm{Jiao}, \binits{L.}},
\bauthor{\bsnm{Qendro}, \binits{L.}},
\bauthor{\bsnm{Kawsar}, \binits{F.}}:
\bctitle{Deepx: A software accelerator for low-power deep learning inference on
  mobile devices}.
In: \bbtitle{2016 15th ACM/IEEE International Conference on Information
  Processing in Sensor Networks (IPSN)},
pp. \bfpage{1}--\blpage{12}
(\byear{2016}).
\bcomment{IEEE}
\end{bchapter}
\endbibitem

%%% 71
\bibitem{zhang2019deep}
\begin{barticle}
\bauthor{\bsnm{Zhang}, \binits{C.}},
\bauthor{\bsnm{Patras}, \binits{P.}},
\bauthor{\bsnm{Haddadi}, \binits{H.}}:
\batitle{Deep learning in mobile and wireless networking: A survey}.
\bjtitle{IEEE Communications surveys \& tutorials}
\bvolume{21}(\bissue{3}),
\bfpage{2224}--\blpage{2287}
(\byear{2019})
\end{barticle}
\endbibitem

%%% 72
\bibitem{ignatov2019ai}
\begin{bchapter}
\bauthor{\bsnm{Ignatov}, \binits{A.}},
\bauthor{\bsnm{Timofte}, \binits{R.}},
\bauthor{\bsnm{Kulik}, \binits{A.}},
\bauthor{\bsnm{Yang}, \binits{S.}},
\bauthor{\bsnm{Wang}, \binits{K.}},
\bauthor{\bsnm{Baum}, \binits{F.}},
\bauthor{\bsnm{Wu}, \binits{M.}},
\bauthor{\bsnm{Xu}, \binits{L.}},
\bauthor{\bsnm{Van~Gool}, \binits{L.}}:
\bctitle{Ai benchmark: All about deep learning on smartphones in 2019}.
In: \bbtitle{2019 IEEE/CVF International Conference on Computer Vision Workshop
  (ICCVW)},
pp. \bfpage{3617}--\blpage{3635}
(\byear{2019}).
\bcomment{IEEE}
\end{bchapter}
\endbibitem

%%% 73
\bibitem{briggs2020federated}
\begin{bchapter}
\bauthor{\bsnm{Briggs}, \binits{C.}},
\bauthor{\bsnm{Fan}, \binits{Z.}},
\bauthor{\bsnm{Andras}, \binits{P.}}:
\bctitle{Federated learning with hierarchical clustering of local updates to
  improve training on non-iid data}.
In: \bbtitle{2020 International Joint Conference on Neural Networks (IJCNN)},
pp. \bfpage{1}--\blpage{9}
(\byear{2020}).
\bcomment{IEEE}
\end{bchapter}
\endbibitem

%%% 74
\bibitem{sztyler2017online}
\begin{bchapter}
\bauthor{\bsnm{Sztyler}, \binits{T.}},
\bauthor{\bsnm{Stuckenschmidt}, \binits{H.}}:
\bctitle{Online personalization of cross-subjects based activity recognition
  models on wearable devices}.
In: \bbtitle{2017 IEEE International Conference on Pervasive Computing and
  Communications (PerCom)},
pp. \bfpage{180}--\blpage{189}
(\byear{2017}).
\bcomment{IEEE}
\end{bchapter}
\endbibitem

\end{thebibliography}
